\documentclass[10pt]{article} %
\usepackage[preprint]{tmlr}

\usepackage{amsmath,amsfonts,bm}

\def\eqref#1{equation~\ref{#1}}

\def\1{\bm{1}}

\def\mI{{\bm{I}}}

\DeclareMathAlphabet{\mathsfit}{\encodingdefault}{\sfdefault}{m}{sl}
\SetMathAlphabet{\mathsfit}{bold}{\encodingdefault}{\sfdefault}{bx}{n}
\newcommand{\tens}[1]{\bm{\mathsfit{#1}}}

\def\tI{{\tens{I}}}

\def\tV{{\tens{V}}}

\def\tX{{\tens{X}}}

\usepackage{hyperref}
\usepackage{url}
\usepackage{graphicx}
\usepackage{wrapfig}
\usepackage{tabularx}
\usepackage{array}
\usepackage{boldline}
\usepackage{algorithmicx}
\usepackage{algorithm}
\usepackage{algpseudocode}
\usepackage{wrapfig}
\usepackage{multirow}
\usepackage{microtype}      
\usepackage{listings}
\usepackage{xcolor}         
\usepackage[table]{xcolor}
\definecolor{Gray}{gray}{0.9}

\usepackage{multirow}
\usepackage{amsmath}
\newcommand{\mycc}{\cellcolor{Gray}}
\usepackage{graphicx}
\newcommand*{\emoticon}[1]{%
  \raisebox{-.2\baselineskip}{%
    \includegraphics[height=1em]{#1}%
  }%
}
\newcommand{\appref}[1]{Appendix~\ref{#1}}
\usepackage{wrapfig}
\usepackage{tabularx}
\usepackage{multicol}
\usepackage{booktabs}
\usepackage[table]{xcolor}
\usepackage{amssymb}    
\usepackage{pifont}     
\usepackage{float}

\title{GriDiT: Factorized Grid-Based Diffusion for Efficient Long Image Sequence Generation}

\author{
\name Snehal Singh Tomar$^{*}$, \name Alexandros Graikos, \name Arjun Krishna, \name Dimitris Samaras, \name Klaus Mueller$^{*}$
\AND
\addr Department of Computer Science\\
  Stony Brook University, Stony Brook, NY\\
  \email $^{*}$Correspondence: \texttt{stomar@cs.stonybrook.edu, mueller@cs.stonybrook.edu}
}

\begin{document}

\maketitle
\begin{abstract}
Modern deep learning methods typically treat image sequences as large tensors of sequentially stacked frames. However, is this straightforward representation ideal given the current state-of-the-art (SoTA)? In this work, we address this question in the context of generative models and aim to devise a more effective way of modeling image sequence data. Observing the inefficiencies and bottlenecks of current SoTA image sequence generation methods, we showcase that rather than working with large tensors, we can improve the generation process by \textit{factorizing} it into first generating the \textit{coarse sequence} at low resolution and then refining the \textit{individual frames} at high resolution. We train a generative model solely on \textit{grid images} comprising subsampled frames. Yet, we learn to generate \textit{image sequences}, using the strong self-attention mechanism of the Diffusion Transformer (DiT) to capture correlations between frames. In effect, our formulation extends a 2D image generator to operate as a low-resolution 3D image-sequence generator without introducing any architectural modifications. Subsequently, we super-resolve each frame individually to add the sequence-independent high-resolution details. This approach offers several advantages and can overcome key limitations of the SoTA in this domain. Compared to existing image sequence generation models, our method achieves superior synthesis quality and improved coherence across sequences. It also delivers high-fidelity generation of arbitrary-length sequences and increased efficiency in inference time and training data usage. Furthermore, our straightforward formulation enables our method to generalize effectively across diverse data domains, which typically require additional priors and supervision to model in a generative context. Our method consistently delivers superior quality and offers a $>2\times$ speedup in inference rates across various datasets. Code is available at our {\color{magenta}\href{https://snehalstomar.github.io/projects/gridit_project/index.html}{project page}}.
\end{abstract}

\section{Introduction}
\label{sec: intro}
\begin{figure}[ht]
    \centering
    \includegraphics[scale = 0.75]{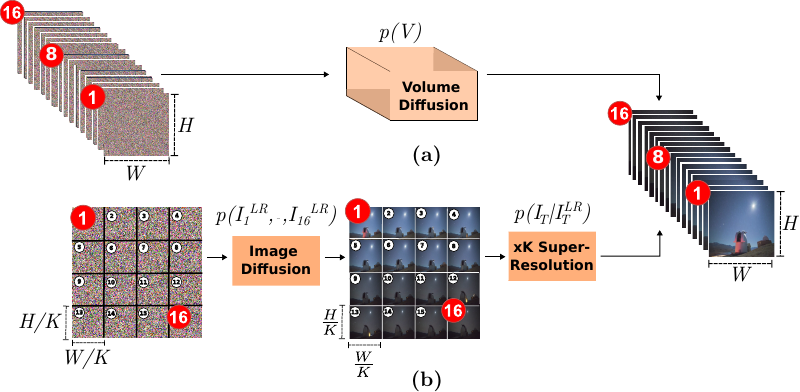}
    \caption{\textbf{(a)} SoTA image sequence generation models treat image sequences as large tensors of ordered frames. In contrast, \textbf{(b)} our method factorizes image sequence generation into two stages. First, we learn to model the dynamics of the sequence at low resolution, treating the frames as subsampled image grids. Second, we learn to super-resolve individual frames at high resolution. Using the DiT's self-attention mechanism to model dynamics across frames, and paired with our sampling strategy, our method yields superior synthesis quality for sequences of arbitrary length while significantly reducing sampling time and training data requirements. (Notation used is the same as defined in section \ref{para: Notation}. $K=4$.)}
    \label{fig: teaser}
\end{figure}
Image sequences form some of the richest perceptual signals in nature, constituting the largest volume of shared data on the internet. At the same time, the complexity of their many degrees of variability (lighting, motion, camera effects, etc.) makes them hard to model. These difficulties have significantly impeded advancements in image sequence generative models, with image sequence generation lagging significantly behind the image and natural language generation paradigms. Only recently has there been an interest \citep{ditctrl_cvpr25, ttt_vid_cvpr25, gs_dit_cvpr_25} in exploring self-attention-driven architectures \citep{txformer, vit, dit} for conditional video generation. However, the benefits of these architectures have yet to be fully exploited by the field.

In the context of this work, we consider \textit{image sequence generation} as an umbrella task encompassing video generation, where the sequential ordering need not necessarily be temporal (e.g., CT scans). Several meaningful attempts have been made to perform the task in the past. However, these attempts fail to bridge long-standing research gaps. 

We envision that the solution to efficient synthesis of arbitrary-length image sequences lies in a well-formed recipe that combines a more effective data representation, an efficient model architecture, and autoregressive sampling. To that effect, we introduce \textbf{GriDiT}: We represent image sequences as grid images comprising subsampled frames, which enables us to factorize image sequence generation into efficient coarse sequence generation at low resolution, followed by refinement of individual frames at high resolution. We pair our generation pipeline with our proposed \textit{Grid-based Autoregressive Sampling Algorithm} to sample sequences of arbitrary length. Figure \ref{fig: teaser} illustrates our approach and contrasts it with prior work. 

Our extensive experiments on the SkyTimelapse \citep{skydset}, CT-RATE \citep{ct-rate-1, ct-rate-2}, Minecraft \citep{mncraft_dset}, and Taichi \citep{taichi_dset} datasets demonstrate that our approach surpasses current state-of-the-art methods in long-range consistency, frame-wise quality, and sampling efficiency for image sequence generation. These advantages become increasingly evident as video length grows. Ablation studies reveal that the core of our improvement lies in coupling Grid-based Modeling with 3D positional embeddings to harness the DiT's \citep{dit} self-attention mechanism effectively. We summarize the key contributions of our work as:
\begin{itemize}
    \item We present a pragmatic outlook towards generating image sequences. We factorize the process into generating coarse sequences at low resolution (Stage 1) and refining individual frames at high resolution (Stage 2). Our approach departs from conventional sequence modeling by treating sequences as image grids, thereby allowing us to use an \textit{image} generation model for \textit{image-sequence} generation. 
    \item We leverage the Diffusion Transformer's self-attention mechanism with our 3D positional embeddings to ensure long-range consistency between the generated frames, achieving superior perceptual quality than the current SoTA on several datasets.
    \item The proposed efficient modeling approach surpasses SoTA methods in sampling time ($>2\times$ faster), training data required (similar performance with 10$\%$ of the training data in data-critical domains), and simplicity of architectures, enabling its applicability to challenging data domains without a domain-specific design.
    \item We facilitate SoTA, arbitrarily long (up to 1024 frames), frame roll out for image sequence generation by introducing a Grid-based Autoregressive sampling algorithm for our diffusion model. 
\end{itemize}

\section{Prior Work}
The image-sequence generation domain has primarily been driven by an \textit{efficiency} versus \textit{synthesis quality} trade-off in recent years. Methods have attempted to make the problem tractable by either modifying model architectures and optimization objectives or utilizing superior embeddings of the sequence. Our position in this work is entirely different from prior art. We present an alternative way of \textit{looking at the data itself} and harness it to tackle the long-existing trade-off effectively.

The high computational cost of processing large video tensors has been a significant impediment in the advancement of image-sequence generation \citep{lvdm, videogpt, MoCoGAN, MoCOGANHD, PVDM, style-gan-v, digan, ddmi, alyosha_long_vid_gan, emu_video, animate_diff}. Most methods \citep{videogpt, MoCoGAN, MoCOGANHD, alyosha_long_vid_gan, videoldm, emu_video} model videos as large tensors, which limits the maximum sequence length and incurs slow inference rates. Recent works using DiTs \citep{ditctrl_cvpr25, gs_dit_cvpr_25, ttt_vid_cvpr25} focus on architectural improvements for better conditioning; we consider these concurrent, but orthogonal to our goal of rethinking image sequence modeling. Approaches leveraging proxy models such as INRs \citep{style-gan-v, digan, ddmi} trade off perceptual quality for efficiency. Factorized generation has shown promise: Emu Video and AnimateDiff \citep{emu_video, animate_diff} split text-to-video into text-to-image and image-to-video stages. In contrast, LongVideoGAN \citep{alyosha_long_vid_gan} factorizes within a GAN-based framework but lacks support for arbitrary-length sequences or resolutions beyond $256{\times}256$. There is a paucity of methods \citep{style-gan-v, lvdm, PVDM, tats} that attempt to generate arbitrary-length videos. Of these, PVDM \citep{PVDM} and LVDM \citep{lvdm} are latent diffusion-based approaches. Whereas TATS \citep{tats} utilizes a GAN, StyleGAN-V is a GAN approach paired with INRs. We compare our method with all relevant techniques to ensure coverage of the various approaches taken to solve the problem. For general image sequence synthesis, we consider the generation of 3D CT volumes. In this regard, GenerateCT \citep{GenCT} is the only method that reports spatio-temporal consistency metrics on publicly available 3D CT data, making it our primary baseline. 

A recent line of work ~\citep{diffusion_forcing, rollling-diffusion} explores alternate noising schemes for the task. 
Of these, we compare with Diffusion Forcing \citep{diffusion_forcing}, which forms a pertinent baseline for comparison as it utilizes autoregressive sampling and its implementation is publicly available. We defer further commentary and contrast with prior art to an exhaustive related work section in \appref{app_sec: rel_work}. To the best of our knowledge, no prior approach in the domain has viewed the problem of image-sequence generation from a standpoint akin to ours.

\section{Our Method}
\label{sec: method}
\begin{figure}
    \centering
    \includegraphics[width=\textwidth]{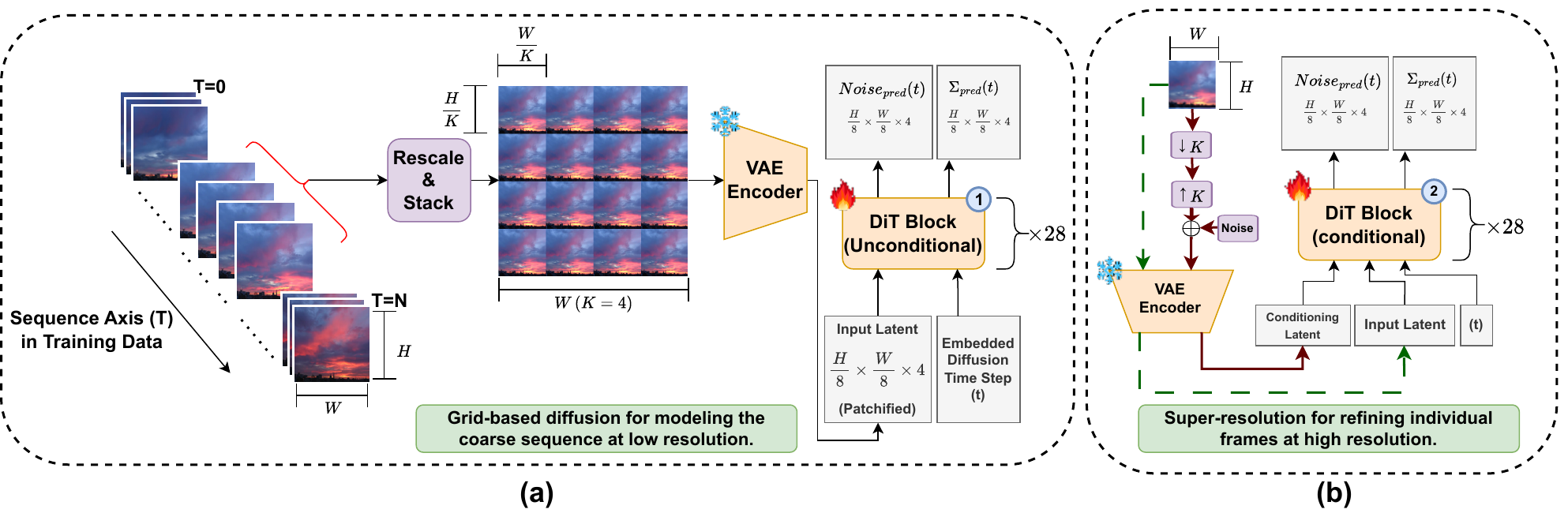}
    \caption{An overview of our method's training pipeline. We leverage DiT's self-attention for efficient, high-quality, and arbitrary-length image sequence generation using a two-stage process. \textbf{(a) Stage 1} (\emoticon{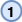}): We learn to generate the \textit{coarse image sequence} at low resolution. We organize the image sequence as grid images, comprising subsampled frames arranged in their sequential order. An unconditional latent DiT is trained to generate them. \textbf{(b) Stage 2} (\emoticon{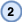}): We learn to refine \textit{individual frames} in the generated coarse sequence via faithful generative $\times K$ super-resolution. We pose the problem as one of learning a conditional DiT model to restore the degradation caused by the lossy subsampling of images from our training dataset. (\emoticon{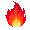}: trainable.   \emoticon{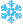}: frozen. $\downarrow K \,\,\, \& \,\,\, \uparrow K$: bilinear (lossy) downsampling and upsampling, respectively. The "Noise" function is further elaborated upon in section \ref{subsec: SR-method}.)}
    \label{fig: training-method}
\end{figure}

\paragraph{Overview.} As shown in Figure \ref{fig: training-method}, we start by modeling 3D image sequences as 2D image grids comprising subsampled frames while preserving their sequential ordering. We then proceed to train an unconditional 
DiT \citep{dit} (Stage 1 (\emoticon{images/one.png})) on these images using the standard DDPM \citep{ddpm} training procedure. At this stage, our model learns to synthesize coarse low-resolution image sequences in the form of 2D grid images. Subsequently, we utilize a conditional DiT-based Super Resolution (SR) (Stage 2 (\emoticon{images/2.png})) pipeline as an up-sampling and refinement mechanism for the generated low-resolution image sequence elements, which we first extract from their respective grids in an order-preserving manner to synthesize high-resolution 3D image sequences. Finally, for inference, we introduce \textit{Grid-based Autoregressive sampling}, which allows us to build on the learned DiT's self-attention mechanism to sample arbitrary-length sequences while only having learned to generate 2D, RGB grid images. We provide a brief preliminary on DDPMs in \appref{app_sec: prelim} for completeness. Both the employed models, viz. Stage-1 and Stage-2 are latent diffusion models trained in the Stable Diffusion \citep{stable_diff} VAE's latent space.  

\paragraph{Notation.} \label{para: Notation} We follow the Notation described here consistently throughout this work. \textbf{Data:} $\tV$ denotes an image sequence and $\tI$ denotes an image in the sequence. $\bar{N}, T, H, W, L,$ and $K$ denote the number of frames in a training image sequence, frame index, frame height, frame width, the total length of the synthetic sequence, and the number of rows and columns in the extracted grid image, respectively. \textbf{Diffusion:} $t,$ 
\begin{wrapfigure}{r}{0.5\textwidth}
    \centering
    \vspace{-1em} 
    \includegraphics[width=\linewidth]{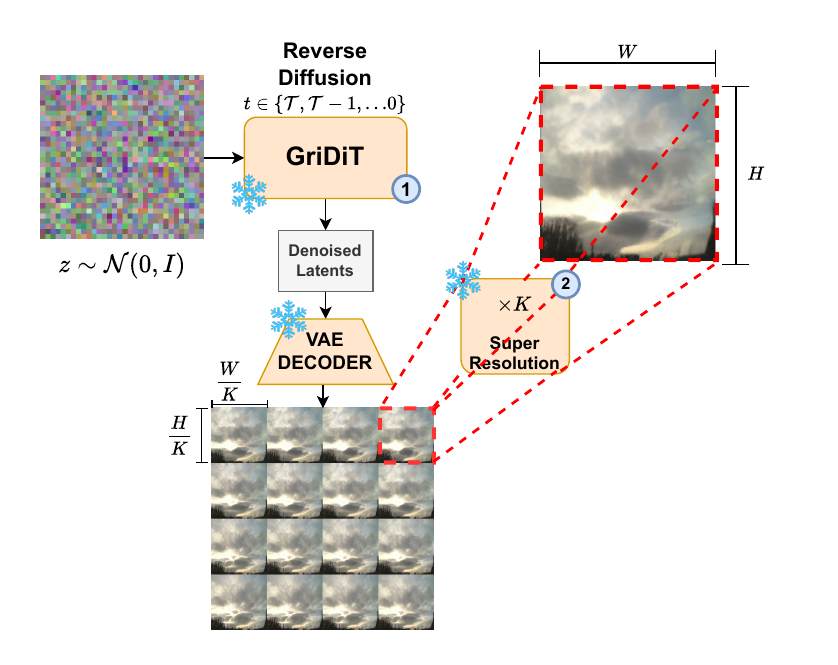}
    \caption{Inferring a single grid image's sequence elements from our model entails: 
    \textbf{(1)} synthesizing grid images using Stage~1 (\emoticon{images/one.png}), 
    \textbf{(2)} splitting the grid into coarse frames, 
    \textbf{(3)} adding fine information and super-resolving the coarse frames into individual output frames via Stage~2 (\emoticon{images/2.png}), and 
    \textbf{(4)} stacking the ordered frames to form the sequence. 
    (\emoticon{images/snowflake.png}: frozen.)}
    \label{fig: inference}
    \vspace{-10mm} 
\end{wrapfigure}
$\tX_{t},$ $\mathcal{T}, \mathcal{T}_{s}, \alpha_t, \bar{\alpha}_t, \epsilon, \epsilon_{\theta}(\tX_{t},t), \beta_{t}$, $\theta$, $\mathcal{N}(0,\mI)$, and $\sigma_{t}$ denote the diffusion timestep, sample at timestep $t$, total number of diffusion timesteps used in training, total number of diffusion timesteps used in sampling, noise mean coefficient, cumulative noise mean coefficient, noise sampled from standard normal distribution, noise predicted by a DiT model, $1-\alpha_t$, model parameters, standard normal distribution, and noise standard deviation coefficient at timestep $t$, respectively. \textbf{Grid-based Autoregressive Sampling:} $T'$ denotes the autoregressive sampling iteration in step 1. $T'' = i[i+1]$ denotes diffusion inpainting-based frame interpolation between the novel frames generated at $T'=i$ and $T'=i+1$ in step 2.
\subsection{Generating Image Sequences with GriDiT}
\label{sec: core_method}
For fixed-length image sequences and videos (or more generally, 2D image volumes) of length $\bar{N}$, a simple video generative model would learn how to sample from the joint distribution of the frames
\begin{equation}
    p(\tV) = p(\tI_{1}, \tI_{2}, \dots, \tI_{T},\dots \tI_{\bar{N}}).
    \label{eq:video_generative_model} 
\end{equation}
In this work, we propose an alternative way (shown in Figure \ref{fig: teaser}) of learning to generate $\tV$ by introducing latent variables that represent the low-resolution frames. Instead of modeling the joint distribution of the high-resolution frames, we model the joint distribution of low-resolution frames, which is cheaper to sample from and can adequately represent the coarse motion between frames. Then, for each frame, we also train a super-resolution module that adds motion-independent 
\begin{figure}[H]
    \centering
    \includegraphics[width=\textwidth]{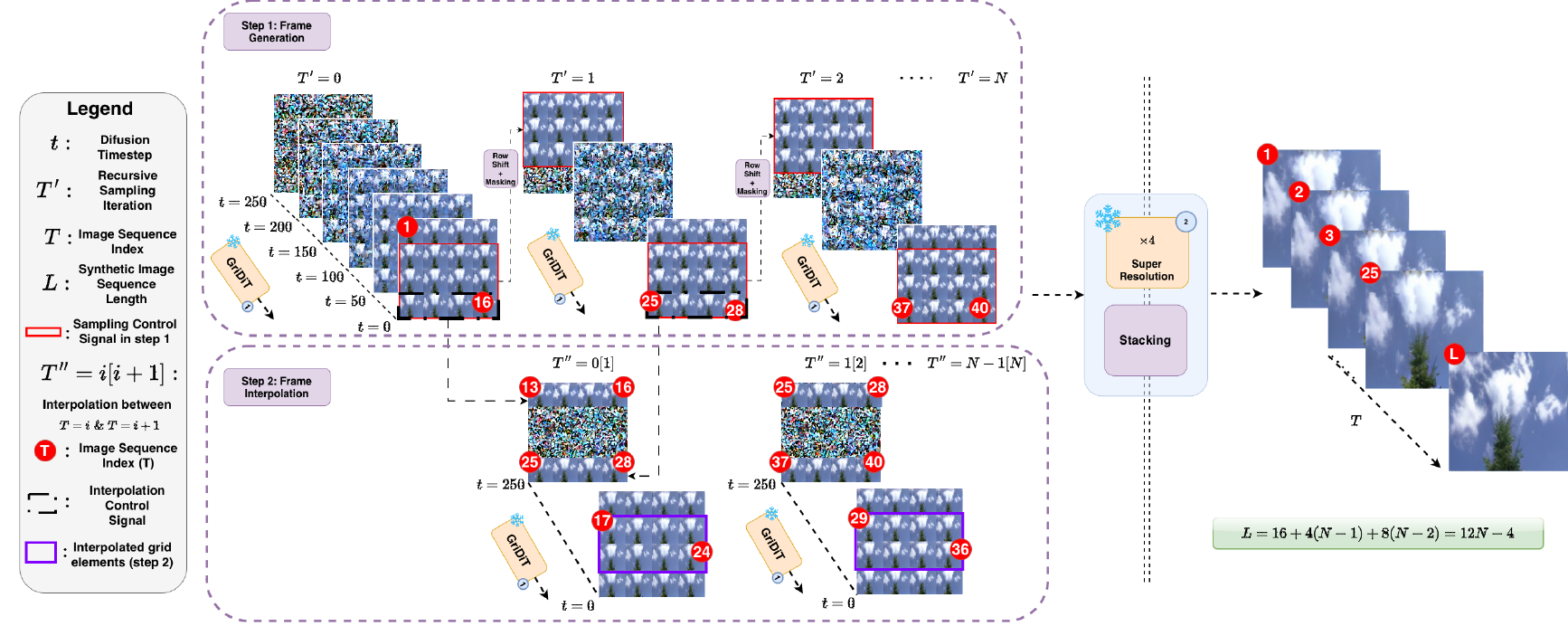}
    \caption{We illustrate our \textbf{Grid-based Autoregressive Sampling Algorithm} used to sample arbitrary-length image sequences. The algorithm entails two steps. We start with \textbf{step 1} wherein the first iteration starts with vanilla Stage 1 (\emoticon{images/one.png}) sampling. Every subsequent iteration uses an appropriately noised control signal from the previous iteration's output at each reverse diffusion timestep to generate four new grid elements, which are in spatiotemporal agreement with all previous grid elements.
    Upon transitioning to \textbf{step 2}, we interpolate eight new frames between each consecutive pair of 4 new frames generated in different sampling iterations of step 1 for superior temporal resolution. Finally, all new coarse grid-elements are super-resolved via Stage 2 (\emoticon{images/2.png}) and stacked in their sequential order for superior spatial resolution and refinement. Consequently, we obtain a long, high-quality image sequence. $N$ such iterations lead to an $L= 12N-4$ length image sequence, inducing a substantial gain in efficiency and quality over the SoTA. (\emoticon{images/snowflake.png}: frozen).}
    \label{fig: grid-recursive-sampling}
\end{figure}
details to refine each low-res frame separately. With this factorization, we disentangle the motion and high-resolution appearance into two separate models, which we find is overall cheaper than modeling the full distribution, while also maintaining high sample quality. Using the previous notation and for $\tI_T^{LR}$ being the low-resolution frames, we can express:

\begin{gather}
    p(\tI_{1}, \ldots, \tI_{\bar{N}},\, \tI_{1}^{LR}, \ldots, \tI_{\bar{N}}^{LR}) =\; p(\tI_{1}, \ldots, \tI_{\bar{N}} \mid \tI_{1}^{LR}, \ldots, \tI_{\bar{N}}^{LR})\,
    p(\tI_{1}^{LR}, \ldots, \tI_{\bar{N}}^{LR}) \nonumber \\[0.5ex] 
    =\; p(\tI_{1} \mid \tI_{1}^{LR}) \cdots p(\tI_{\bar{N}} \mid \tI_{\bar{N}}^{LR})\;
     \,p(\tI_{1}^{LR}, \ldots, \tI_{\bar{N}}^{LR}),
    \label{eq:our_generative_model}
\end{gather}

Where $p(\tI_1^{LR},\dots,\tI_{\bar{N}}^{LR})$ models the joint distribution of the low-res frames represented as grid images and $p(\tI_T \mid \tI_T^{LR})$ super-resolves each frame. We remark that $ p(\tI_1, \ldots, \tI_{\bar{N}},\, \tI_1^{LR}, \ldots, \tI_{\bar{N}}^{LR})$ is easily tractable given $p(\tV)$ (Eq. \ref{eq:video_generative_model}) because $\tI_T^{LR}$ can be obtained from $\tI_T$ by bicubic downsampling. We parameterize learning the learned distribution of low resolution frames represented by grid images as $p_{\theta_1}(\tI_1^{\textit{LR}},\ldots,\tI_{\bar{N}}^{\textit{LR}})$, and denote the learned optimal stage 1 (\emoticon{images/one.png})  parameters as $\theta_1^{\ast}$. Similarly, we parameterize learning the refinement of coarse individual frames as $p_{\theta_2}(\tI_T|\tI_T^{\textit{LR}})$, and denote the learned optimal stage 2 (\emoticon{images/2.png}) parameters as $\theta_2^{\ast}$. We elucidate our model architectures in Section \ref{app_sec: model_arch} of the Appendix. 

Here, we emphasize that learning to model $p(\tI_1^{LR}, \ldots, \tI_{\bar{N}}^{LR})$ with GriDiT is significantly less compute-intensive than learning to model $p(\tV)$ (Eq~\ref{eq:video_generative_model}) via SoTA methods. Hence, this formulation is crucial for the performance of our method, as highlighted in our experiments.

\subsection{Grid-based Frame Modeling}
\label{subsec: grid-based-frame-modeling}
To model sequences as grids of images comprising subsampled frames, we arrange a subsequence of $K^{2}$ frames, subsampled by a factor $K$, into a grid per their sequential ordering to form an image representing the frames (or slices) of the video (or volume) data, at a lower spatial resolution. The process is illustrated in Figure \ref{fig: training-method} and further expanded formally in \appref{app_sec: grid-frame-modeling}. Here, we remark that our grid-image formulation does not alter the temporal ordering in the image sequences in any way. Instead, it simply \textit{rearranges} the (subsampled) frames per their original temporal order. The obtained grid image tensor is now suitable for training the image diffusion model. We employ an unconditional DiT model (Stage 1 (\emoticon{images/one.png})) \citep{dit} to learn to sample from the distribution of grid images $p_{\theta_{1}}(\tI_1^{LR}, \ldots, \tI_{\bar{N}}^{LR})$. We use $K=4$ in most experiments.
\paragraph{3D Positional Embeddings.} 
\label{subsec: meth-3D-pos-embed}
The DiT model uses fixed 2D positional embeddings to inform each patch of its spatial location. Since we are modeling multiple frames in a single image, each pixel should not only be aware of its spatial neighbors, but also of the neighboring pixels in different frames. We employ 3D positional embeddings to encode the cross-frame locality. This is a straightforward extension of the 3D positional embeddings used in video transformer models \citep{vivit}. We first compute the 3D positional embeddings on the downsampled image volume and rearrange them into the grid format to combine with the 2D embeddings. We present a formal description of our positional embeddings in \appref{app_sec: 3d_pos_embed}.

\paragraph{Grid-based Autoregressive Sampling.} 
We posit that Autoregressive sampling holds the key to generating high-quality, arbitrary-length image sequences. To that end, we introduce Grid-based Autoregressive sampling to sample arbitrary-length image sequences from our unconditional DiT (Stage 1 (\emoticon{images/one.png})) model, which is trained solely to generate 2D RGB grid images. These grid images comprise $K^{2}$ length sub-sequences in the form of subsampled grid elements. The two-step approach, illustrated by Figure \ref{fig: grid-recursive-sampling}, draws inspiration from diffusion-based image inpainting literature \citep{repaint}. Diffusion inpainting \textit{fills} "missing" segments of an image that are coherent with the "known" segments by modifying the reverse-diffusion (sampling) process. The modification substitutes the regions corresponding to the "known" segments within the denoised latent (starting from $\mathcal{N}(0,I)$ at $t=\mathcal{T}$) at each reverse-sampling timestep $t$ with their appropriately forward-noised variants (at $t$) obtained from the ground truth. Thereby allowing the diffusion process and the denoiser model's priors to \textit{inpaint} "missing" information in accordance with the "known" information. Conditional diffusion inpainting of "missing" grid elements, in accordance with "known" grid elements, forms the foundation of our sampling approach. In our case, the diffusion process ensures spatial coherence, and the Stage 1 (\emoticon{images/one.png})  model's learned implicit temporal bias from grid images ensures temporal coherence.    

In Step 1 of our algorithm, we generate a coarse sequence of grid elements that maintain spatiotemporal coherence. We begin with vanilla DDPM sampling to generate the start-of-sequence grid image ($K^2$ new grid elements). Each subsequent iteration $T'$ results in the generation of 4 new coarse frames that bear spatiotemporal coherence with the 12 immediately previous frames. The coherence is ensured via conditional diffusion inpainting, with the last three rows of the prior iteration, $T'-1$, serving as the sampling control signal. The coarse frames obtained here act as inputs to step 2.

Step 2 of our algorithm focuses on enhancing the temporal resolution of the sequence. At each iteration $T'' = [i] i+1$, We interpolate eight new grid elements between the novel grid elements synthesized at $T'=i$ and $T'= i+1$. We do so by using the latest synthetic row from $T'=i$ as the first row and that from $T'=i+1$ as the last row in our conditional diffusion inpainting framework. 

Finally, we split and stack all newly created grid elements according to their intended temporal order. The resulting coarse frames are now suitable for spatial super-resolution and refinement using our Stage 2 model. Upon conclusion, $N$ stage 1 sampling iterations appropriately paired with $N-1$ step 2 iterations lead to a synthetic sequence of length $L= 12N-4$. A formal overview of our sampling technique, along with algorithmic summaries of Steps 1 and 2, can be found in \appref{app_sec:grid-based-ar-sampling}. 
 
\subsection{Frame Refinement by Generative Super-Resolution (SR)}
\label{subsec: SR-method}
We use super-resolution to refine individual frames from the coarse sequence generated via Grid-based Autoregressive sampling from the learned distribution $p_{\theta_1^{\ast}}(\tI_1^{\textit{LR}},\ldots,\tI_{\bar{N}}^{\textit{LR}})$. We specifically chose a diffusion 
\begin{figure}[H]
    \centering
    \includegraphics[width=\textwidth]{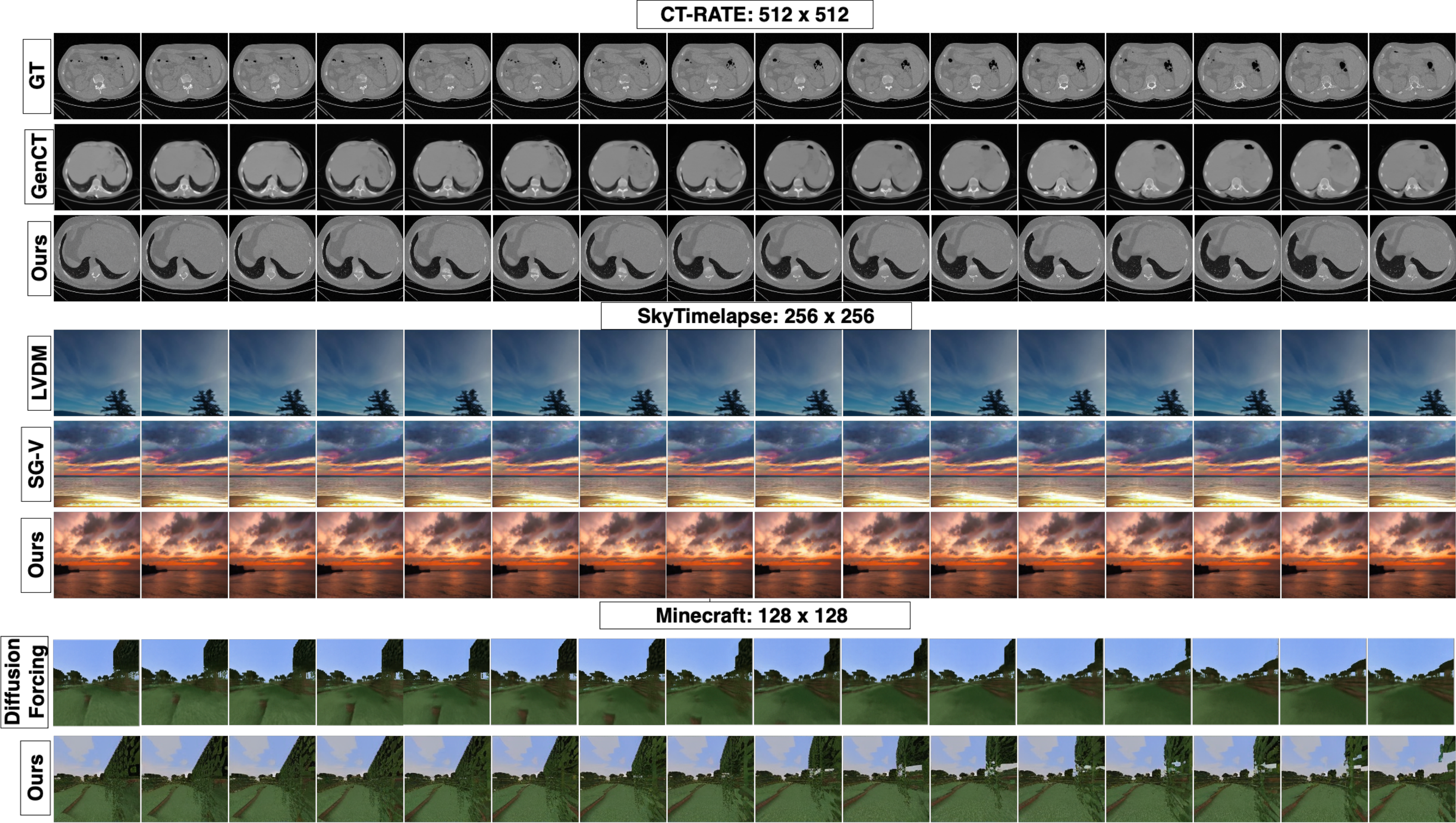}
    \caption{\textbf{Qualitative comparisons} with the SoTA on the CT-RATE SkyTimeLapse, and Minecraft datasets. Images are arranged from left to right in their sequential order, i.e., frames 1 through 16. Synthetic CT volumes are generated at 512 $\times$ 512 resolution using $\times$4 SR whereas SkyTimelapse videos are generated at 256 $\times$ 256 resolution using $\times$2 SR. We use the standard $4\times4$ grid setting in Grid-based Autoregressive sampling (step-1) for both cases. Whereas we use a $\{8\times8$ grid, four-row control signal$\}$ setting for step 1 of sampling in experiments on the Minecraft dataset to intentionally allow room for $\times2$ SR, ensuring a fair comparison while yielding the desired 128 $\times$ 128 resolution. Our method yields superior performance in terms of spatiotemporal coherence and quality. (GT: ground truth, GenCT: GenerateCT, SG-V:  StyleGAN-V)}
    \label{fig: qual-results}
\end{figure}
\begin{table}[ht]
\begin{minipage}[t]{0.52\textwidth}
    \centering
    \caption{3D CT Volume synthesis benchmark.\\\textbf{Bold:} best. \textsuperscript{\textdagger}: numbers reported by GenerateCT .}
    \label{tab:main-quant-ct}
    \setlength{\tabcolsep}{4pt}
    \small
    \resizebox{0.99\textwidth}{!}{
    \begin{tabular}{lcccc}
    \toprule
    \textbf{Method} & \textbf{Data} \% & \textbf{FVD} \textdownarrow & \textbf{FID} \textdownarrow & \textbf{Time (s)} \textdownarrow  \\
    \midrule
    Base w/ Imagen & 100 & 3557.7\textsuperscript{\textdagger} & 160.8\textsuperscript{\textdagger} & 234 \\
    Base w/ SD & 100 & 3513.5\textsuperscript{\textdagger} & 151.7\textsuperscript{\textdagger} & 367 \\
    Base w/ Phenaki & 100 & 1886.8\textsuperscript{\textdagger} & 104.3\textsuperscript{\textdagger} & 197 \\
    GenCT & 100 & 1092.3\textsuperscript{\textdagger} & 55.8\textsuperscript{\textdagger} & 184 \\
    \rowcolor{Gray}
    Ours & 10 & 1089.5 & 68.2 & \textbf{53.8} \\
    \rowcolor{Gray}
    Ours & 60 & 1079.6 & 61.5 & \textbf{53.8} \\
    \rowcolor{Gray}
    \textbf{Ours} & 100 & \textbf{998.43} & \textbf{54.8} & \textbf{53.8} \\
    \bottomrule
    \end{tabular}
    }
    \label{tab:main-results}
\end{minipage}
\hfill
\begin{minipage}[t]{0.47\textwidth}
    \centering
    \small
    \setlength{\tabcolsep}{4pt}
    \renewcommand{\arraystretch}{1.2}
    \caption{Quantitative comparisons on the Minecraft dataset. \textbf{Bold:} best.}
    \resizebox{\textwidth}{!}{%
    \begin{tabular}{l|cccc}
    \toprule
     & \multicolumn{4}{c}{\textbf{Number of Synthetic Frames}} \\
    \cmidrule(lr){2-5}
    \textbf{Method} & \textbf{16} & \textbf{128} & \textbf{256} & \textbf{1024} \\
    \midrule
     & \multicolumn{4}{c}{\textbf{FVD} $\downarrow$} \\
    \cmidrule(lr){2-5}
    \multicolumn{1}{l|}{\textbf{Diffusion Forcing}} & 62.43 & 199.117 & 221.53 & 261.23 \\
    \rowcolor{gray!15} \textbf{Ours} & 64.32 & \textbf{184.728} & \textbf{218.69} & \textbf{243.21} \\
    \midrule
     & \multicolumn{4}{c}{\textbf{Flicker (Average $l_{1}$ distance between frames) $\downarrow$}} \\
    \cmidrule(lr){2-5}
    \multicolumn{1}{l|}{\textbf{Diffusion Forcing}} & \textbf{20.42} & \textbf{21.99} & 27.63 & 32.13 \\
    \rowcolor{gray!15} \textbf{Ours} & 20.79 & 22.31 & \textbf{25.68} & \textbf{31.86} \\
    \bottomrule
    \end{tabular}}
    \label{tab:fvd_flicker}
\end{minipage}%
\end{table}
model to perform SR, which we denote as $p_{\theta_2^{\ast}}(\tI_T|\tI_T^{\textit{LR}})$, to hallucinate some details that are lost in the low-resolution images. Vanilla DiT does not support direct image conditioning. Therefore, we make specific modifications in our Stage 2 (\emoticon{images/2.png}) architecture. As illustrated in Figure \ref{fig: training-method} (b), we train stage-2 to super-resolve the low-resolution frames using a conditional DiT model. In this case, we utilize the DiT model with certain modifications, as we are generating a single, high-resolution (HR) image conditioned on its low-resolution (LR) counterpart. More specifically, we obtain LR images from our training datasets of HR images by applying a combination of degradations (successive lossy down and upsampling, noise addition, and blurring) to the HR images. We elaborate on the employed degradation scheme, providing experimental justification for it in \appref{app_sec: custom_sr_justification}. Our training dataset for the task now comprises several $\{$LR, HR$\}$ pairs.  

For each pair, the SR model's goal is to learn to generate HR images conditioned on the corresponding LR input. We train it to do so by embedding both LR and HR images in the VAE's \citep{vae} latent space, concatenating the two latents,  projecting the concatenated latent on the original hidden dimension, and training the DiT to generate the embedding for HR given the obtained projected embedding as input. We apply conditioning to the DiT via adaptive layer norm. We summarize the process of Stage-2 inference, or going from a coarse synthetic grid image comprising subsampled frames to highly photorealistic and motion preserving individual frames, in Figure \ref{fig: inference}.

\section{Experiments}
\label{sec: exp}
\subsection{Setup} 

\paragraph{Datasets.} We make use of the Skytimelapse \citep{skydset}, Taichi \citep{taichi_dset}, and Minecraft \citep{mncraft_dset} datasets at $256^2$, $256^2$, and $128^2$ resolution, respectively, for evaluating our method on the arbitrary length video generation task. We utilize the CT-RATE dataset \citep{ct-rate-1, ct-rate-2} at a resolution of $512^2$ for evaluations on the 3D CT Volume generation task. 

\paragraph{Experimental Specifics.} We use $K=4, \text{and} \,\,\, \mathcal{T}_{s} = 250$ in all our experimental results except wherever we specify otherwise. We defer other design choices with respect to training and experimentation to \appref{app_sec: exp_setup}

\paragraph{Baselines.} We construct relevant baselines with widely benchmarked prior works to compare against our model's performance. Our baselines encompass both GAN and diffusion-based methods for completeness. In the context of video generation: (1) On the SkyTimelapse dataset we compare with VideoGPT \citep{videogpt}, MoCoGAN \citep{MoCoGAN}, MoCoGAN-HD \citep{MoCOGANHD}, LVDM \citep{lvdm}, PVDM \citep{PVDM}, DIGAN \citep{digan}, StyleGAN-V \citep{style-gan-v}, and DDMI \citep{ddmi} for standard (length: 16 an 128 frames) and with StyleGAN-V \citep{style-gan-v} and LVDM \citep{lvdm} for arbitrarily long generation, respectively. Our choice of competing methods is derived from StyleGAN-V \citep{style-gan-v} and DDMI \citep{ddmi}; (2) On the Taichi dataset, we compare with  LVDM \citep{lvdm}, TATS \citep{tats}, DIGAN \citep{digan}, and Style-SV \citep{stylesv} in different length settings as appropriate for each method. This choice of baselines is dictated by various other relevant methods in the domain. For image sequence (3D CT Volume) generation on the CT-RATE dataset, we borrow our baselines from GenerateCT \citep{GenCT}, which were formed by appropriately finetuning Stable Diffusion (Base w/ SD) \citep{stable_diff}, Imagen (Base w/ Imagen) \citep{Imagen}, and Phenaki (Base w/ Phenaki) \citep{phenaki}. We note that Phenaki is a video-generation model, and comparing it ensures completeness with respect to the types of chosen competing methods. We compare with only the best variants of competing methods. For the sake of fairness in comparison, we borrowed metrics reported in prior work and utilized publicly available pre-trained weights wherever feasible. We retrain certain modules of other methods if they are fit for comparison on the experiment in question and do not report weights or provide pre-trained weights. More specifically, we retrained relevant modules of LVDM and Style-SV on the Taichi dataset. In \appref{app_sec:concurr_work}, we present our concurrent works and define the scope of our comparative analysis.
\begin{figure}
    \centering
\includegraphics[width=\linewidth]{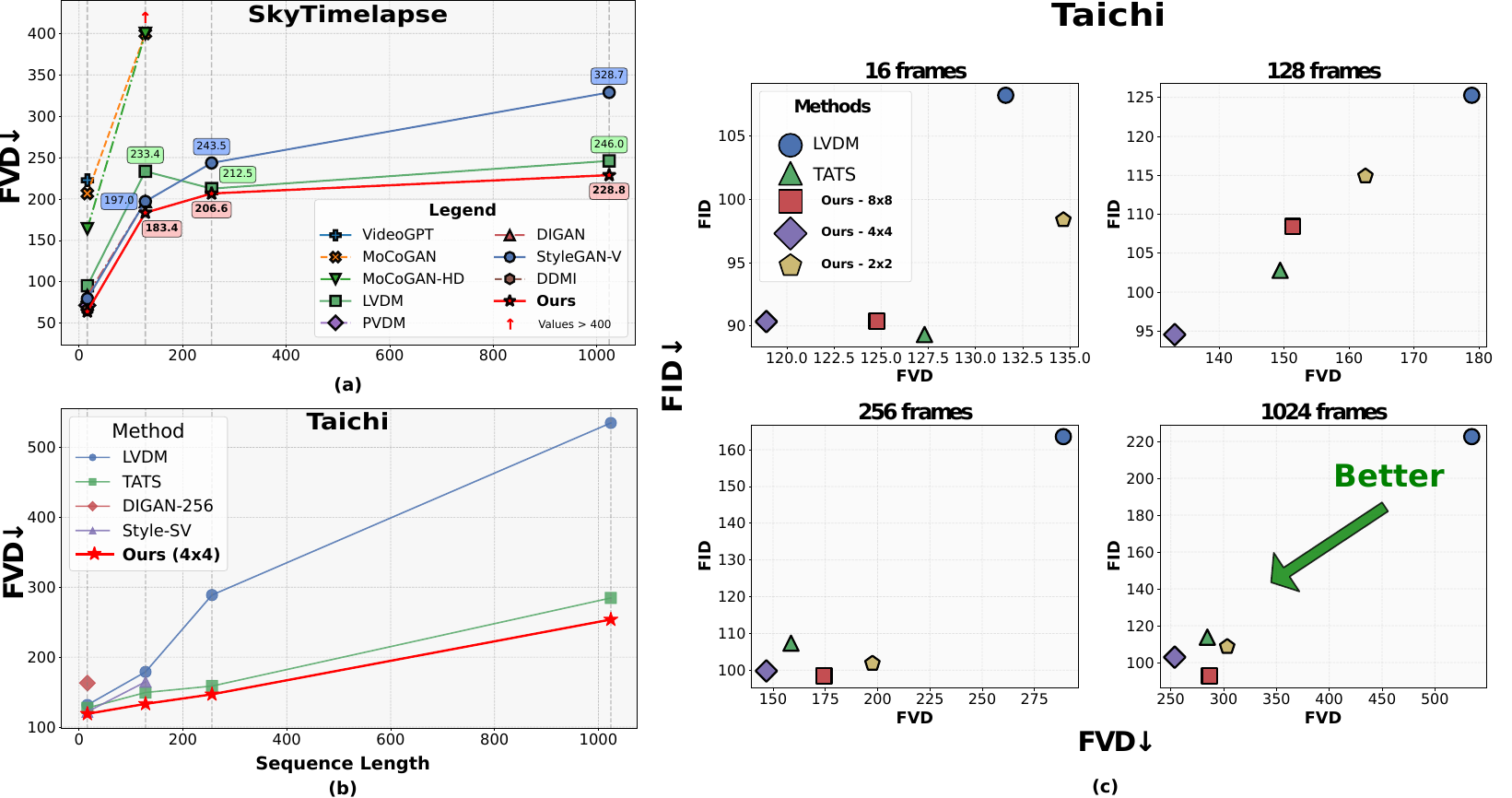}
    \caption{Quantitative comparisons with the SoTA for arbitrary length generation on the \textbf{(a)} SkyTimeLapse \textbf{(b)} Taichi datasets. We observe that our method outperforms the SoTA convincingly, and our advantage over the SoTA increases monotonically with increasing sequence length. \textbf{(c)} We study the effect of varying the grid size ($K$) on the observed FVD and FID on synthetic Taichi videos.}
    \label{fig: main-quant}
\end{figure}
\begin{table}[ht]
\centering
\caption{We present a consolidated analysis of the \textbf{sampling time (s)} achieved by our method and its variants in different video length settings on the SkyTimelapse and Taichi datasets at $256\times256$ resolution. Our method massively outperforms the SoTA in all settings and the comparative advantage grows with increasing video length. We report the highlighted rows as our method's benchmark considering all associated tradeoffs. (\textbf{Bold}: best entry. underline: second best entry.)}
\label{tab: sampling-time-long}
\resizebox{\textwidth}{!}{%
\begin{tabular}{lcccccccccc}
\toprule
&\multicolumn{4}{c}{\textbf{Video Length (SkyTimelapse)}} && &\multicolumn{4}{c}{\textbf{Video Length (Taichi)}} \\
\cmidrule(lr){2-5} \cmidrule(lr){8-11}
\textbf{Method} & 16 & 128 & 256 & 1024 & & \textbf{Method} & 16 & 128 & 256 & 1024 \\
\midrule
VideoGPT      & 58.7   & -      & -       & -      & & LVDM        & 91.7   & 273.4   & 284.8   & 1204.1 \\
MoCoGAN-HD    & 77.8    & -      & -       & -      & & TATS        & 95.8   & 287.5  & 482.6    & 1308.4 \\
LVDM          & 63.7   & \underline{112.8} & \underline{245.4}   & \underline{1186.1} & & Style-SV    & 86.5   & 241.4  & -        & - \\
PVDM          & \underline{47.6}    & 168.0    & -       & -      & & Ours - 2x2    & \underline{42.3}   & 206.8  & 307.2  & 1380.0 \\
StyleGAN-V    & 62.0      & 243.2 & 386.3  & 1467.2 & & \mycc\textbf{Ours - 4x4}    & \mycc\textbf{4.8}     & \mycc\underline{38.27}   & \mycc\underline{79.66}  & \mycc\underline{515.43} \\
\mycc \textbf{Ours} & \mycc \textbf{4.8} & \mycc \textbf{38.27} & \mycc \textbf{79.66} & \mycc \textbf{515.43} & & Ours - 8x8    & \textbf{4.8}     & \textbf{15.0}  & \textbf{26.5}   & \textbf{51.4} \\
\bottomrule
\end{tabular}
}
\end{table}
\definecolor{lightblue}{rgb}{0.9,0.95,1.0}
\definecolor{lightgray}{gray}{0.95}
\definecolor{lightyellow}{rgb}{1.0,1.0,0.9}

\begin{table}[h]
\centering
\caption{We observe that our 3D positional embeddings yield superior perceptual quality (FVD, FID) and sptiotemporal consistency (FVD) than conventional 2D positional embeddings across all experimental setups and sequence lengths. (\textdownarrow : lower is better. \textbf{Bold}: best entry.)}
\label{tab:pos-embed-ablation}
\resizebox{\textwidth}{!}{%
{\small
\begin{tabular}{lcc>{\columncolor{lightblue}}c>{\columncolor{lightblue}}c>{\columncolor{lightgray}}c>{\columncolor{lightgray}}c>{\columncolor{lightyellow}}c>{\columncolor{lightyellow}}c}
\toprule
\textbf{Setup} & \textbf{Resolution} & \textbf{Pos Emb.} & \textbf{FVD-16} \textdownarrow & \textbf{FID-16} \textdownarrow & \textbf{FVD-128} \textdownarrow & \textbf{FID-128} \textdownarrow & \textbf{FVD-201} \textdownarrow & \textbf{FID-201} \textdownarrow \\
\midrule
CT-RATE & 512 & 2D & 289.74 & 58.8  & 378.2  & 56.7   & 1074.8  & 55.6  \\
CT-RATE & 512 & 3D & \textbf{268.3} & \textbf{53.3}  & \textbf{356.1}  & \textbf{55.2}   & \textbf{998.9}  & \textbf{54.8}  \\
\midrule
SkyTimelapse & 256 & 2D & 71.2    & 41.3   & 199.6  & 44.8   & -       & -     \\
SkyTimelapse & 256 & 3D & \textbf{64.1}   & \textbf{37.7}   & \textbf{183.4}   & \textbf{40.9}  & -       & -     \\
\midrule
Taichi (4$\times$4) & 256      & 2D & 139.8  & 159.5 & 152.6  & 167.6  & -       & -     \\
Taichi (4$\times$4)  & 256     & 3D & \textbf{118.9}  & \textbf{157.3} & \textbf{133.2} & \textbf{164.6}& -       & -     \\
\bottomrule
\end{tabular}
}
}
\end{table}
\subsection{Our Results}
\label{subsec: results}
\paragraph{Synthesis quality of arbitrary length sequences.} Conventional fixed-length image sequence synthesis models treat videos as stacked frame tensors, inherently limiting sequence-length flexibility. Consequently, they perform qualitative assessments typically only up to 128 frames. In contrast, in Figure \ref{fig: main-quant} we compare our method with the SoTA on the (a) SkyTimelapse and (b)  Taichi datasets for sequence generation up to 1024 frames at $256\times256$ resolution. We also present the data shown in Figure \ref{fig: main-quant} (a) in tabular form in \appref{app_sec: sky_quant} for the 16 and 128 length video settings for visual comfort. On CT-RATE, we compare with GenerateCT, a specialized 3D CT generator that utilizes a three-stage architecture, CT priors, and language guidance for generating 3D CT volumes (comprising 201 slices) at a resolution of $512\times512$ in Table \ref{tab:main-quant-ct}. We utilized our Grid-based Autoregressive sampling to generate all the sequences used in our comparisons. The results demonstrate the following: \textbf{(1)} We conclusively outperform prior approaches comprising different modeling paradigms, viz. GANs, INRs, Diffusion models, and their combinations. \textbf{(2)} Our approach is domain-agnostic and does not require additional priors or supervision to model unconventional data. \textbf{(3)} Our design elements come together to support faithful arbitrary-length general image sequence synthesis. \textbf{(4)} Despite training only on sequences having $\le400$ frames, our method generates much longer high-fidelity sequences. Moreover, our relative superiority over other methods becomes more pronounced as the length of sampled sequences increases. Thereby confirming robust generalization, free of leakage \citep{leakage} or memorization artifacts \citep{memorization}.

Figure \ref{fig: qual-results} shows qualitative comparisons with the SoTA for standard video generation. For GenerateCT we used the prompt `44 years old male: The overall examination is within normal limits' to sample. All other sequences shown in the figure were sampled unconditionally. Our method exhibits improved sequence consistency and sharpness in CT volumes compared to the SoTA. GenerateCT's volumes show random jumps (e.g., frames $12^{th}$ to $13^{th}$ and $15^{th}$ to $16^{th}$). On SkyTimeLapse, LVDM produces blurry videos with low variability, while StyleGAN-V generates unrealistic lighting. Our approach avoids these issues, producing more coherent and realistic samples.
Since we model motion at low resolution, we must investigate the potential artifacts that could emerge due to insufficient modeling resolution and potentially lossy super-resolution. To that end, we compare our method with Diffusion Forcing \citep{diffusion_forcing} on the Minecraft dataset \citep{mncraft_dset}. The reasons for our experimental choice are twofold. First, the Minecraft dataset comprises gameplay videos that contain large amounts of motion content per frame. Second, Diffusion Forcing belongs to a recent class of literature that intervenes with the Diffusion noise scheme in Autoregressive video generation. Consequently, comparing with it ensures completeness in our evaluation. Here, we remark that although Minecraft videos warrant a $128^{2}$ resolution only, we chose our $\{8\times8$ grid, four row control signal$\}$ setting for step 1 of sampling to intentionally allow room for $\times2$ SR, ensuring fair comparison. We report these experimental results in Figure \ref{fig: qual-results} and Table \ref{tab:fvd_flicker}. Therein, we employ the average $l_1$ distance between consecutive frames as a metric for flicker, following ~\citet{cogvideox} who use it in the same context. We found that our method performs comparably to Diffusion Forcing at shorter synthetic sequence lengths and outperforms it at longer sequence lengths. The trend is consistent both in terms of synthesis quality and flicker. Thereby establishing that our method's limitation of modeling motion only at low resolution does not become a handicap even when generating sequences with large amounts of motion.      
\paragraph{Synthesis Efficiency.} \label{par: efficiency} Our method is more efficient than previous methods in: (1) \textbf{inference speed};  As reported in Tables \ref{tab: sampling-time-long} and \ref{tab:main-quant-ct}, our model is consistently $>2\times$ faster than SoTA across all three data domains and across variable sequence dimensions. We do observe, however, that our advantage over the SoTA reduces with increasing sequence length due to iterative SR that is slower. (2) \textbf{training data required}; we investigate this property in Table \ref{tab:main-quant-ct} wherein we observe that our method attains superior FVD and comparable slice-wise FID scores when compared with GenerateCT on the CT-RATE dataset with as little as $10\%$ of the training data. Moreover, the performance improves monotonically with increasing data. (3) \textbf{simplicity}; in the case of the CT-RATE dataset, our approach is significantly simpler in contrast to GenerateCT, which employs a three-stage CT-prior dependent approach, requiring text conditioning to achieve the reported quality metrics.

\subsection{Ablation Studies}
\label{subsec: ablations}
\paragraph{Grid Size $(K)$.} In Figure \ref{fig: main-quant} (c), we analyze the effect of varying $K \in \{2,4,8\}$ in training and step 1 of sampling on the attained synthesis quality for videos sampled via \textit {one, three, and four} rows as control signals in step 1 and interpolating half the rows in step 2 of sampling via the Grid-based Autoregressive Sampling algorithm. We conducted this experiment using different image sequence lengths on the Taichi dataset, which is challenging due to significant motion across frames and fine-grained details. We observe that (1) there 
\begin{wrapfigure}{r}{0.5\textwidth}
  \centering
  \includegraphics[width=1\linewidth]{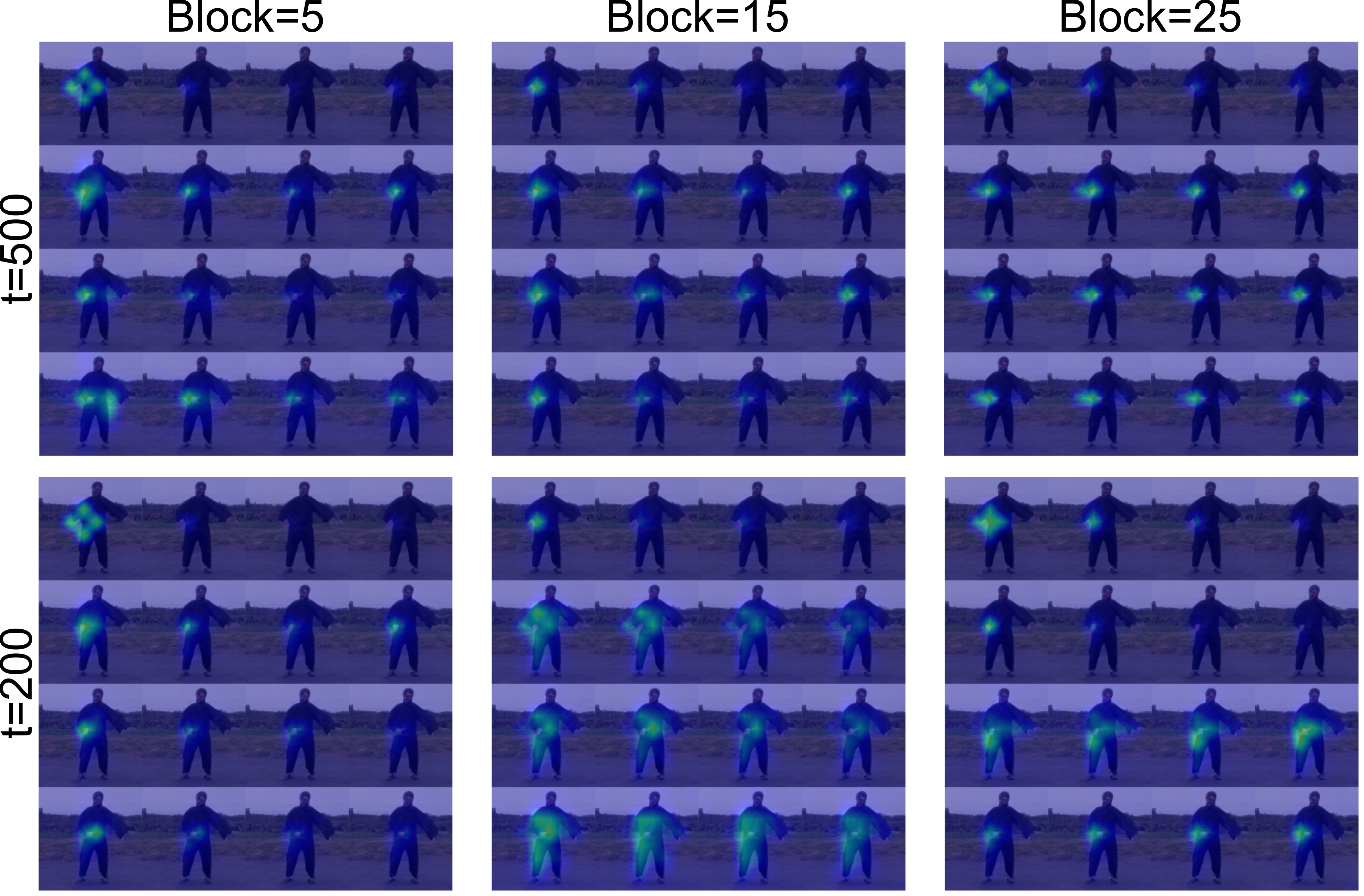} 
  \caption{We visualize our Stage 1 (\emoticon{images/one.png}) model's $1024\times1024$ attention maps scaled and overlayed on a corresponding sampled Taichi grid image. The emergence of high attention scores in grid-like patterns, bearing a direct correlation with the synthetic grid elements, suggests a strong self-attention prior that is key to our generation and sampling pipeline. ($t:$ reverse diffusion timestep. Block: $r$ denotes the $r^{th}$ DiT block.)}
  \label{fig: attention-analysis}
\vspace{-20pt}
\end{wrapfigure}
exists a tradeoff between the amount of long-range sequence modeling signal and frame-wise fine information that a grid size setting has to offer. For instance, $K=8$ offers a higher temporal span for DiT's self-attention mechanism but causes significant loss of finer details in the subsampled grid elements. Whereas $K=2$ is limited in the time field, it does not cause any loss of high-frequency information, as $ H=512$, $ W=512$, and the required frame resolution is $256$. Despite that, $K=8$ offers superior FVD and FID over $K=2$ for all sequence lengths. Thereby, establishing that superior sequential modeling is more instrumental than resolution preservation for sequence generation with our method. (2) Our method gains in sequence modeling while sacrificing a little on the quality of individual frames. The same is reflected in the FVD versus FID tradeoff we observe here. (3) The setting $K=4$ sits at a sweet spot between both tradeoffs, consequently, making it our setting of choice for obtaining most results.
\paragraph{Positional Embedding.} We observed certain `looping artifacts' when generating sequences with our method using the vanilla DiT. In that, the frames would move back and forth in terms of motion, rather than having smooth transitions. We posit that the behavior was caused by the DiT's use of 2D positional embeddings, which are not suitable for consistently modeling motion across frames (subsampled grid elements in our case). To that end, we used 3D positional embeddings in our method as outlined in section \ref{subsec: meth-3D-pos-embed}. In Table \ref{tab:pos-embed-ablation}, we justify our choice of 3D positional embeddings, demonstrating that they consistently benefit our method across all settings. We also observe the same result qualitatively, with complete remediation of the `looping artifacts'.
We present additional details in \appref{app_sec: 3d_pos_embed}.

\paragraph{Role of refinement via SR.} We investigate the contribution of our SR (Stage-2 \emoticon{images/2.png}) model to the overall synthesis quality achieved by our method by contrasting the performance of our trained SR models with that of a SoTA off-the-shelf SR method, SinSR \citep{sinsr}. We observe that our stage-2 outperforms SinSR in all settings, most importantly on the CT-RATE dataset, suggesting it is necessary to finetune the SR method in out-of-distribution settings (e.g., super-resolving medical images). We provide more details in \appref{app_sec: custom_sr_justification}.

\paragraph{Mechanistic Insights.} The DiT's strong self-attention mechanism is crucial to our approach since the generation of coherent grid images and the Grid-based Autoregressive Sampling algorithm both heavily rely on it. Consequently, we visualize attention maps from our Stage-1 (\emoticon{images/one.png}) model, scaled and overlaid onto synthetic grid images in Figure \ref{fig: attention-analysis} to gain a better understanding of the attention mechanism in our method’s context. Therein, we observe that high attention scores emerge in grid-like patterns, suggesting that regions within grid elements attend mostly to corresponding regions within other grid elements. This thereby forms the basis of the observed sequence-wise consistency and provides evidence for the soundness of our grid-based formulation. 

We defer additional experimental results to the appendices. Specifically, we explore our model's ability to function as a plug-and-play image restoration model in \appref{app_sec: denoising}, where we perform 3D CT volume denoising using our method. Our diffusion-based approach outperforms previous baselines on this task. We also provide additional experimental evidence supporting the unique positioning of our approach within the image sequence generation landscape in \appref{app_sec:positioning}. Finally, we include additional qualitative results in \appref{appsec:additional_qual}.

\section{Limitations and Future Work}
\label{sec: limits_future}
By modeling sequences as grids of subsampled images, we incur certain losses in capturing the fine motion between frames. Even though these losses are insignificant on most real-world datasets, as shown in our experiments, they could potentially lead to inconsistencies for unobservable moving objects at low resolution. Along similar lines, the interpolation involved in our sampling scheme may cause undesired smoothening for video datasets captured at significantly higher frame rates than most real-world datasets. Additionally, our autoregressive long-form video generation relies on naive diffusion inpainting, which could be improved with more efficient algorithms. We have not been able to study the scaling laws for our models presented in this work due to computational constraints. It would be interesting to scale our models to the multi-billion parameter size group and compare their performance with relevant baselines currently excluded from our scope of analysis.
We further elaborate upon the societal implications of our work in our detailed impact statement in \appref{app_sec: impacts}
Addressing these limitations, incorporating multi-modal conditioning, and ethics and safety studies constitute a promising design space for future work.
\section{Conclusion}
Image sequences have historically been treated as large tensors of stacked frames. As a result, fixed-length synthesis, subpar sequential coherence, and prohibitively slow sampling rates have been some of the long-standing limitations in their generation. We consider them as grid images comprising subsampled frames instead, which are later super-resolved back to the original resolution. Therefore, being able to effectively harness modern self-attention based architectures and autoregressive sampling for the task whilst gaining on efficiency. Our method offers superior synthesis quality, efficiency, and support for generating arbitrary-length sequences without relying on proxy approximators, such as INRs. It generalizes well to specialized domains, such as 3D CT volumes, without prior-driven designs. The presented ablation studies show that GriDiT's underlying mechanisms conform to its theoretical formulation and substantiate our design choices. Overall, GriDiT establishes a strong framework for scalable image-sequence generation and opens avenues for improved pixel-space data representations in future research.

\bibliography{main}

@String(CVPR= {IEEE Conf. Comput. Vis. Pattern Recog.})

@String(ICCV= {Int. Conf. Comput. Vis.})

@String(ECCV= {Eur. Conf. Comput. Vis.})

@String(ICLR = {Int. Conf. Learn. Represent.})

@String(CVPR  = {CVPR})

@String(ICCV  = {ICCV})

@String(ECCV  = {ECCV})

@String(ICLR  = {ICLR})

@inproceedings{txformer,
 author = {Vaswani, Ashish and Shazeer, Noam and Parmar, Niki and Uszkoreit, Jakob and Jones, Llion and Gomez, Aidan N and Kaiser, \L ukasz and Polosukhin, Illia},
 booktitle = {Advances in Neural Information Processing Systems},
 editor = {I. Guyon and U. Von Luxburg and S. Bengio and H. Wallach and R. Fergus and S. Vishwanathan and R. Garnett},
 pages = {},
 publisher = {Curran Associates, Inc.},
 title = {Attention is All you Need},
 url = {https://proceedings.neurips.cc/paper_files/paper/2017/file/3f5ee243547dee91fbd053c1c4a845aa-Paper.pdf},
 volume = {30},
 year = {2017}
}

@misc{videogpt,
      title={VideoGPT: Video Generation using VQ-VAE and Transformers}, 
      author={Wilson Yan and Yunzhi Zhang and Pieter Abbeel and Aravind Srinivas},
      year={2021},
      eprint={2104.10157},
      archivePrefix={arXiv},
      primaryClass={cs.CV}
}

@inproceedings{
vit,
title={An Image is Worth 16x16 Words: Transformers for Image Recognition at Scale},
author={Alexey Dosovitskiy and Lucas Beyer and Alexander Kolesnikov and Dirk Weissenborn and Xiaohua Zhai and Thomas Unterthiner and Mostafa Dehghani and Matthias Minderer and Georg Heigold and Sylvain Gelly and Jakob Uszkoreit and Neil Houlsby},
booktitle={International Conference on Learning Representations},
year={2021},
url={https://openreview.net/forum?id=YicbFdNTTy}
}

@inproceedings{MoCoGAN,
  title={{MoCoGAN}: Decomposing motion and content for video generation},
  author={Tulyakov, Sergey and Liu, Ming-Yu and Yang, Xiaodong and Kautz, Jan},
  booktitle={IEEE Conference on Computer Vision and Pattern Recognition (CVPR)},
  pages={1526--1535},
  year={2018}
}

@inproceedings{MoCOGANHD,
  title={A Good Image Generator Is What You Need for High-Resolution Video Synthesis},
  author={Tian, Yu and Ren, Jian and Chai, Menglei and Olszewski, Kyle and Peng, Xi and Metaxas, Dimitris N. and Tulyakov, Sergey},
  booktitle={International Conference on Learning Representations (ICLR)},
  year={2021}
}

@inproceedings{digan,
  title={Generating Videos with Dynamics-aware Implicit Generative Adversarial Networks},
  author={Yu, Sihyun and Tack, Jihoon and Mo, Sangwoo and Kim, Hyunsu and Kim, Junho and Ha, Jung-Woo and Shin, Jinwoo},
  booktitle={International Conference on Learning Representations},
  year={2022}
}

@inproceedings{PVDM,
  title={Video Probabilistic Diffusion Models in Projected Latent Space},
  author={Yu, Sihyun and Sohn, Kihyuk and Kim, Subin and Shin, Jinwoo},
  booktitle={Proceedings of the IEEE/CVF Conference on Computer Vision and Pattern Recognition},
  year={2023}
}

@article{lvdm,
  title={Latent video diffusion models for high-fidelity long video generation},
  author={He, Yingqing and Yang, Tianyu and Zhang, Yong and Shan, Ying and Chen, Qifeng},
  journal={arXiv preprint arXiv:2211.13221},
  year={2022}
}

@inproceedings{ddmi,
  title={DDMI: Domain-Agnostic Latent Diffusion Models for Synthesizing High-Quality Implicit Neural Representations},
  author={Park, Dogyun and Kim, Sihyeon and Lee, Sojin and Kim, Hyunwoo J},
  booktitle={The Twelfth International Conference on Learning Representations},
  year={2024}
}

@inproceedings{videoldm,
    title={Align your Latents: High-Resolution Video Synthesis with Latent Diffusion Models},
    author={Blattmann, Andreas and Rombach, Robin and Ling, Huan and Dockhorn, Tim and Kim, Seung Wook and Fidler, Sanja and Kreis, Karsten},
    booktitle={IEEE Conference on Computer Vision and Pattern Recognition ({CVPR})},
    year={2023}
}

@inproceedings{skydset,
  title={DTVNet: Dynamic time-lapse video generation via single still image},
  author={Zhang, Jiangning and Xu, Chao and Liu, Liang and Wang, Mengmeng and Wu, Xia and Liu, Yong and Jiang, Yunliang},
  booktitle={European Conference on Computer Vision},
  pages={300--315},
  year={2020},
  organization={Springer}
}

@inproceedings{
warmup,
title={{SGDR}: Stochastic Gradient Descent with Warm Restarts},
author={Ilya Loshchilov and Frank Hutter},
booktitle={International Conference on Learning Representations},
year={2017},
url={https://openreview.net/forum?id=Skq89Scxx}
}

@InProceedings{GenCT,
author="Hamamci, Ibrahim Ethem
and Er, Sezgin
and Sekuboyina, Anjany
and Simsar, Enis
and Tezcan, Alperen
and Simsek, Ayse Gulnihan
and Esirgun, Sevval Nil
and Almas, Furkan
and Do{\u{g}}an, Irem
and Dasdelen, Muhammed Furkan
and Prabhakar, Chinmay
and Reynaud, Hadrien
and Pati, Sarthak
and Bluethgen, Christian
and Ozdemir, Mehmet Kemal
and Menze, Bjoern",
editor="Leonardis, Ale{\v{s}}
and Ricci, Elisa
and Roth, Stefan
and Russakovsky, Olga
and Sattler, Torsten
and Varol, G{\"u}l",
title="GenerateCT: Text-Conditional Generation of 3D Chest CT Volumes",
booktitle="Computer Vision -- ECCV 2024",
year="2025",
publisher="Springer Nature Switzerland",
address="Cham",
pages="126--143",
isbn="978-3-031-72986-7"
}

@inproceedings{
Imagen,
title={Photorealistic Text-to-Image Diffusion Models with Deep Language Understanding},
author={Chitwan Saharia and William Chan and Saurabh Saxena and Lala Li and Jay Whang and Emily Denton and Seyed Kamyar Seyed Ghasemipour and Raphael Gontijo-Lopes and Burcu Karagol Ayan and Tim Salimans and Jonathan Ho and David J. Fleet and Mohammad Norouzi},
booktitle={Advances in Neural Information Processing Systems},
editor={Alice H. Oh and Alekh Agarwal and Danielle Belgrave and Kyunghyun Cho},
year={2022},
url={https://openreview.net/forum?id=08Yk-n5l2Al}
}

@inproceedings{
phenaki,
title={Phenaki: Variable Length Video Generation from Open Domain Textual Descriptions},
author={Ruben Villegas and Mohammad Babaeizadeh and Pieter-Jan Kindermans and Hernan Moraldo and Han Zhang and Mohammad Taghi Saffar and Santiago Castro and Julius Kunze and Dumitru Erhan},
booktitle={International Conference on Learning Representations},
year={2023},
url={https://openreview.net/forum?id=vOEXS39nOF}
}

@misc{fvd,
title={{FVD}: A new Metric for Video Generation},
author={Thomas Unterthiner and Sjoerd van Steenkiste and Karol Kurach and Rapha{\"e}l Marinier and Marcin Michalski and Sylvain Gelly},
year={2019},
url={https://openreview.net/forum?id=rylgEULtdN}
}

@InProceedings{style-gan-v,
    author    = {Skorokhodov, Ivan and Tulyakov, Sergey and Elhoseiny, Mohamed},
    title     = {StyleGAN-V: A Continuous Video Generator With the Price, Image Quality and Perks of StyleGAN2},
    booktitle = {Proceedings of the IEEE/CVF Conference on Computer Vision and Pattern Recognition (CVPR)},
    month     = {June},
    year      = {2022},
    pages     = {3626-3636}
}

@InProceedings{i3d,
author = {Carreira, Joao and Zisserman, Andrew},
title = {Quo Vadis, Action Recognition? A New Model and the Kinetics Dataset},
booktitle = {Proceedings of the IEEE Conference on Computer Vision and Pattern Recognition (CVPR)},
month = {July},
year = {2017}
}

@InProceedings{dit,
    author    = {Peebles, William and Xie, Saining},
    title     = {Scalable Diffusion Models with Transformers},
    booktitle = {Proceedings of the IEEE/CVF International Conference on Computer Vision (ICCV)},
    month     = {October},
    year      = {2023},
    pages     = {4195-4205}
}

@misc{vae,
      title={Auto-Encoding Variational Bayes}, 
      author={Diederik P Kingma and Max Welling},
      year={2022},
      eprint={1312.6114},
      archivePrefix={arXiv},
      primaryClass={stat.ML},
      url={https://arxiv.org/abs/1312.6114}, 
}

@InProceedings{stable_diff,
    author    = {Rombach, Robin and Blattmann, Andreas and Lorenz, Dominik and Esser, Patrick and Ommer, Bj\"orn},
    title     = {High-Resolution Image Synthesis With Latent Diffusion Models},
    booktitle = {Proceedings of the IEEE/CVF Conference on Computer Vision and Pattern Recognition (CVPR)},
    month     = {June},
    year      = {2022},
    pages     = {10684-10695}
}

@inproceedings{sinsr,
  title={SinSR: diffusion-based image super-resolution in a single step},
  author={Wang, Yufei and Yang, Wenhan and Chen, Xinyuan and Wang, Yaohui and Guo, Lanqing and Chau, Lap-Pui and Liu, Ziwei and Qiao, Yu and Kot, Alex C and Wen, Bihan},
  booktitle={Proceedings of the IEEE/CVF Conference on Computer Vision and Pattern Recognition},
  pages={25796--25805},
  year={2024}
}

@article{ct-rate-1,
  title={A foundation model utilizing chest CT volumes and radiology reports for supervised-level zero-shot detection of abnormalities},
  author={Hamamci, Ibrahim Ethem and Er, Sezgin and Almas, Furkan and Simsek, Ayse Gulnihan and Esirgun, Sevval Nil and Dogan, Irem and Dasdelen, Muhammed Furkan and Wittmann, Bastian and Simsar, Enis and Simsar, Mehmet and others},
  journal={arXiv preprint arXiv:2403.17834},
  year={2024}
}

@article{ct-rate-2,
  title={Ct2rep: Automated radiology report generation for 3d medical imaging},
  author={Hamamci, Ibrahim Ethem and Er, Sezgin and Menze, Bjoern},
  journal={arXiv preprint arXiv:2403.06801},
  year={2024}
}

@InProceedings{repaint,
    author    = {Lugmayr, Andreas and Danelljan, Martin and Romero, Andres and Yu, Fisher and Timofte, Radu and Van Gool, Luc},
    title     = {RePaint: Inpainting Using Denoising Diffusion Probabilistic Models},
    booktitle = {Proceedings of the IEEE/CVF Conference on Computer Vision and Pattern Recognition (CVPR)},
    month     = {June},
    year      = {2022},
    pages     = {11461-11471}
}

@article{ddpm,
  title={Denoising diffusion probabilistic models},
  author={Ho, Jonathan and Jain, Ajay and Abbeel, Pieter},
  journal={Advances in neural information processing systems},
  volume={33},
  pages={6840--6851},
  year={2020}
}

@inproceedings{imagenet,
  title={Imagenet: A large-scale hierarchical image database},
  author={Deng, Jia and Dong, Wei and Socher, Richard and Li, Li-Jia and Li, Kai and Fei-Fei, Li},
  booktitle={2009 IEEE conference on computer vision and pattern recognition},
  pages={248--255},
  year={2009},
  organization={Ieee}
}

@article{score-1,
  author  = {Aapo Hyv{{\"a}}rinen},
  title   = {Estimation of Non-Normalized Statistical Models by Score Matching},
  journal = {Journal of Machine Learning Research},
  year    = {2005},
  volume  = {6},
  number  = {24},
  pages   = {695-709},
  url     = {http://jmlr.org/papers/v6/hyvarinen05a.html}
}

@inproceedings{score-2,
 author = {Song, Yang and Ermon, Stefano},
 booktitle = {Advances in Neural Information Processing Systems},
 editor = {H. Wallach and H. Larochelle and A. Beygelzimer and F. d\textquotesingle Alch\'{e}-Buc and E. Fox and R. Garnett},
 pages = {},
 publisher = {Curran Associates, Inc.},
 title = {Generative Modeling by Estimating Gradients of the Data Distribution},
 url = {https://proceedings.neurips.cc/paper_files/paper/2019/file/3001ef257407d5a371a96dcd947c7d93-Paper.pdf},
 volume = {32},
 year = {2019}
}

@InProceedings{unets,
author="Ronneberger, Olaf
and Fischer, Philipp
and Brox, Thomas",
editor="Navab, Nassir
and Hornegger, Joachim
and Wells, William M.
and Frangi, Alejandro F.",
title="U-Net: Convolutional Networks for Biomedical Image Segmentation",
booktitle="Medical Image Computing and Computer-Assisted Intervention -- MICCAI 2015",
year="2015",
publisher="Springer International Publishing",
address="Cham",
pages="234--241",
isbn="978-3-319-24574-4"
}

@InProceedings{miccai_3d_ct,
author="Jeong, Jiheon
and Kim, Ki Duk
and Nam, Yujin
and Cho, Kyungjin
and Kang, Jiseon
and Hong, Gil-Sun
and Kim, Namkug",
editor="Greenspan, Hayit
and Madabhushi, Anant
and Mousavi, Parvin
and Salcudean, Septimiu
and Duncan, James
and Syeda-Mahmood, Tanveer
and Taylor, Russell",
title="Generating High-Resolution 3D CT with 12-Bit Depth Using a Diffusion Model with Adjacent Slice and Intensity Calibration Network",
booktitle="Medical Image Computing and Computer Assisted Intervention -- MICCAI 2023",
year="2023",
publisher="Springer Nature Switzerland",
address="Cham",
pages="366--375",
isbn="978-3-031-43999-5"
}

@InProceedings{tats,
author="Ge, Songwei
and Hayes, Thomas
and Yang, Harry
and Yin, Xi
and Pang, Guan
and Jacobs, David
and Huang, Jia-Bin
and Parikh, Devi",
editor="Avidan, Shai
and Brostow, Gabriel
and Ciss{\'e}, Moustapha
and Farinella, Giovanni Maria
and Hassner, Tal",
title="Long Video Generation with Time-Agnostic VQGAN and Time-Sensitive Transformer",
booktitle="Computer Vision -- ECCV 2022",
year="2022",
publisher="Springer Nature Switzerland",
address="Cham",
pages="102--118",
isbn="978-3-031-19790-1"
}

@inproceedings{alyosha_long_vid_gan,
 author = {Brooks, Tim and Hellsten, Janne and Aittala, Miika and Wang, Ting-Chun and Aila, Timo and Lehtinen, Jaakko and Liu, Ming-Yu and Efros, Alexei and Karras, Tero},
 booktitle = {Advances in Neural Information Processing Systems},
 editor = {S. Koyejo and S. Mohamed and A. Agarwal and D. Belgrave and K. Cho and A. Oh},
 pages = {31769--31781},
 publisher = {Curran Associates, Inc.},
 title = {Generating Long Videos of Dynamic Scenes},
 url = {https://proceedings.neurips.cc/paper_files/paper/2022/file/ce208d95d020b023cba9e64031db2584-Paper-Conference.pdf},
 volume = {35},
 year = {2022}
}

@InProceedings{emu_video,
author="Girdhar, Rohit
and Singh, Mannat
and Brown, Andrew
and Duval, Quentin
and Azadi, Samaneh
and Rambhatla, Sai Saketh
and Shah, Akbar
and Yin, Xi
and Parikh, Devi
and Misra, Ishan",
editor="Leonardis, Ale{\v{s}}
and Ricci, Elisa
and Roth, Stefan
and Russakovsky, Olga
and Sattler, Torsten
and Varol, G{\"u}l",
title="Factorizing Text-to-Video Generation by Explicit Image Conditioning",
booktitle="Computer Vision -- ECCV 2024",
year="2025",
publisher="Springer Nature Switzerland",
address="Cham",
pages="205--224",
isbn="978-3-031-73033-7"
}

@article{animate_diff,
  title={AnimateDiff: Animate Your Personalized Text-to-Image Diffusion Models without Specific Tuning},
  author={Guo, Yuwei and Yang, Ceyuan and Rao, Anyi and Liang, Zhengyang and Wang, Yaohui and Qiao, Yu and Agrawala, Maneesh and Lin, Dahua and Dai, Bo},
  journal={International Conference on Learning Representations},
  year={2024}
}

@article{gpt,
  author       = {Tom B. Brown and
                  Benjamin Mann and
                  Nick Ryder and
                  Melanie Subbiah and
                  Jared Kaplan and
                  Prafulla Dhariwal and
                  Arvind Neelakantan and
                  Pranav Shyam and
                  Girish Sastry and
                  Amanda Askell and
                  Sandhini Agarwal and
                  Ariel Herbert{-}Voss and
                  Gretchen Krueger and
                  Tom Henighan and
                  Rewon Child and
                  Aditya Ramesh and
                  Daniel M. Ziegler and
                  Jeffrey Wu and
                  Clemens Winter and
                  Christopher Hesse and
                  Mark Chen and
                  Eric Sigler and
                  Mateusz Litwin and
                  Scott Gray and
                  Benjamin Chess and
                  Jack Clark and
                  Christopher Berner and
                  Sam McCandlish and
                  Alec Radford and
                  Ilya Sutskever and
                  Dario Amodei},
  title        = {Language Models are Few-Shot Learners},
  journal      = {CoRR},
  volume       = {abs/2005.14165},
  year         = {2020},
  url          = {https://arxiv.org/abs/2005.14165},
  eprinttype    = {arXiv},
  eprint       = {2005.14165},
  bibsource    = {dblp computer science bibliography, https://dblp.org}
}

@inproceedings{xlnet,
  title={XLNet: Generalized Autoregressive Pretraining for Language Understanding},
  author={Yang, Zhilin and Dai, Zihang and Yang, Yiming and Carbonell, Jaime and Salakhutdinov, Ruslan and Le, Quoc V.},
  booktitle={Advances in Neural Information Processing Systems (NeurIPS)},
  year={2019},
  url={https://papers.nips.cc/paper/2019/hash/dc6a7e655d7e5840e66733e9ee67cc69-Abstract.html}
}

@inproceedings{nova,
  title={Autoregressive Video Generation without Vector Quantization},
  author={Deng, Haoge and Pan, Ting and Diao, Haiwen and Luo, Zhengxiong and Cui, Yufeng and Lu, Huchuan and Shan, Shiguang and Qi, Yonggang and Wang, Xinlong},
  booktitle={International Conference on Learning Representations (ICLR)},
  year={2025},
  url={https://openreview.net/forum?id=JE9tCwe3lp}
}

@inproceedings{magvitv2,
  title={Language Model Beats Diffusion -- Tokenizer is Key to Visual Generation},
  author={Yan, Wilson and Yin, Xinyuan and Liu, Jeff and Xu, Shang and Liu, Xuanlin and Lin, Richard and Petrov, Mikhail and Darrell, Trevor and Huang, Jiajun and Stone, Peter and others},
  booktitle={International Conference on Learning Representations (ICLR)},
  year={2024},
  url={https://arxiv.org/abs/2310.05737}
}

@inproceedings{
diffusionblend_ct,
title={DiffusionBlend: Learning 3D Image Prior through Position-aware Diffusion Score Blending for 3D Computed Tomography Reconstruction},
author={Bowen Song and Jason Hu and Zhaoxu Luo and Jeffrey A Fessler and Liyue Shen},
booktitle={The Thirty-eighth Annual Conference on Neural Information Processing Systems},
year={2024},
url={https://openreview.net/forum?id=h3Kv6sdTWO}
}

@inproceedings{
ddgsct_ct,
title={{DDGS}-{CT}: Direction-Disentangled Gaussian Splatting for Realistic Volume Rendering},
author={Zhongpai Gao and Benjamin Planche and Meng Zheng and Xiao Chen and Terrence Chen and Ziyan Wu},
booktitle={The Thirty-eighth Annual Conference on Neural Information Processing Systems},
year={2024},
url={https://openreview.net/forum?id=mY0ZnS2s9u}
}

@inproceedings{
lung_ct,
title={Lung250M-4B: A Combined 3D Dataset for {CT}- and Point Cloud-Based Intra-Patient Lung Registration},
author={Fenja Falta and Christoph Gro{\ss}br{\"o}hmer and Alessa Hering and Alexander Bigalke and Mattias P Heinrich},
booktitle={Thirty-seventh Conference on Neural Information Processing Systems Datasets and Benchmarks Track},
year={2023},
url={https://openreview.net/forum?id=FC0dsvguFi}
}

@InProceedings{brain_mri,
    author    = {Young, Sean I. and Balbastre, Yael and Fischl, Bruce and Golland, Polina and Iglesias, Juan Eugenio},
    title     = {Fully Convolutional Slice-to-Volume Reconstruction for Single-Stack MRI},
    booktitle = {Proceedings of the IEEE/CVF Conference on Computer Vision and Pattern Recognition (CVPR)},
    month     = {June},
    year      = {2024},
    pages     = {11535-11545}
}

@INPROCEEDINGS{vivit,
  author={Arnab, Anurag and Dehghani, Mostafa and Heigold, Georg and Sun, Chen and Lučić, Mario and Schmid, Cordelia},
  booktitle={2021 IEEE/CVF International Conference on Computer Vision (ICCV)}, 
  title={ViViT: A Video Vision Transformer}, 
  year={2021},
  volume={},
  number={},
  pages={6816-6826},
  doi={10.1109/ICCV48922.2021.00676}}

@InProceedings{taichi_dset,
  author={Siarohin, Aliaksandr and Lathuilière, Stéphane and Tulyakov, Sergey and Ricci, Elisa and Sebe, Nicu},
  title={First Order Motion Model for Image Animation},
  booktitle = {Conference on Neural Information Processing Systems (NeurIPS)},
  month = {December},
  year = {2019}
}

@article{big_batchsz,title	= {Measuring the Effects of Data Parallelism on Neural Network Training},author	= {Chris Shallue and Jaehoon Lee and Joseph Antognini and Jascha Sohl-dickstein and Roy Frostig and George Dahl},year	= {2018},URL	= {https://arxiv.org/pdf/1811.03600.pdf},journal	= {Journal of Machine Learning Research (JMLR)}}

@inproceedings{
fp-16,
title={Mixed Precision Training},
author={Paulius Micikevicius and Sharan Narang and Jonah Alben and Gregory Diamos and Erich Elsen and David Garcia and Boris Ginsburg and Michael Houston and Oleksii Kuchaiev and Ganesh Venkatesh and Hao Wu},
booktitle={International Conference on Learning Representations},
year={2018},
url={https://openreview.net/forum?id=r1gs9JgRZ},
}

@INPROCEEDINGS {grad_acc,
author = { Andersson, Axel and Koriakina, Nadezhda and Sladoje, Natasa and Lindblad, Joakim },
booktitle = { 2022 IEEE International Conference on Big Data (Big Data) },
title = {{ End-to-end Multiple Instance Learning with Gradient Accumulation }},
year = {2022},
volume = {},
ISSN = {},
pages = {2742-2746},
doi = {10.1109/BigData55660.2022.10020801},
url = {https://doi.ieeecomputersociety.org/10.1109/BigData55660.2022.10020801},
publisher = {IEEE Computer Society},
address = {Los Alamitos, CA, USA},
month =Dec}

@inproceedings{fid,
 author = {Heusel, Martin and Ramsauer, Hubert and Unterthiner, Thomas and Nessler, Bernhard and Hochreiter, Sepp},
 booktitle = {Advances in Neural Information Processing Systems},
 editor = {I. Guyon and U. Von Luxburg and S. Bengio and H. Wallach and R. Fergus and S. Vishwanathan and R. Garnett},
 pages = {},
 publisher = {Curran Associates, Inc.},
 title = {GANs Trained by a Two Time-Scale Update Rule Converge to a Local Nash Equilibrium},
 url = {https://proceedings.neurips.cc/paper_files/paper/2017/file/8a1d694707eb0fefe65871369074926d-Paper.pdf},
 volume = {30},
 year = {2017}
}

@article{stylesv,
  title={Towards Smooth Video Composition},
  author={Zhang, Qihang and Yang, Ceyuan and Shen, Yujun and Xu, Yinghao and Zhou, Bolei},
  journal={International Conference on Learning Representations (ICLR)},
  year={2023}
}

@inproceedings{memorization,
 author = {van den Burg, Gerrit and Williams, Chris},
 booktitle = {Advances in Neural Information Processing Systems},
 editor = {M. Ranzato and A. Beygelzimer and Y. Dauphin and P.S. Liang and J. Wortman Vaughan},
 pages = {27916--27928},
 publisher = {Curran Associates, Inc.},
 title = {On Memorization in Probabilistic Deep Generative Models},
 url = {https://proceedings.neurips.cc/paper_files/paper/2021/file/eae15aabaa768ae4a5993a8a4f4fa6e4-Paper.pdf},
 volume = {34},
 year = {2021}
}

@InProceedings{leakage,
    author    = {Somepalli, Gowthami and Singla, Vasu and Goldblum, Micah and Geiping, Jonas and Goldstein, Tom},
    title     = {Diffusion Art or Digital Forgery? Investigating Data Replication in Diffusion Models},
    booktitle = {Proceedings of the IEEE/CVF Conference on Computer Vision and Pattern Recognition (CVPR)},
    month     = {June},
    year      = {2023},
    pages     = {6048-6058}
}

@article{ditctrl_cvpr25,
  title     = {DiTCtrl: Exploring Attention Control in Multi-Modal Diffusion Transformer for Tuning-Free Multi-Prompt Longer Video Generation},
  author    = {Cai, Minghong and Cun, Xiaodong and Li, Xiaoyu and Liu, Wenze and Zhang, Zhaoyang and Zhang, Yong and Shan, Ying and Yue, Xiangyu},
  journal   = {arXiv:2412.18597},
  year      = {2024},
}

@misc{ttt_vid_cvpr25,
      title={One-Minute Video Generation with Test-Time Training}, 
      author={Karan Dalal and Daniel Koceja and Gashon Hussein and Jiarui Xu and Yue Zhao and Youjin Song and Shihao Han and Ka Chun Cheung and Jan Kautz and Carlos Guestrin and Tatsunori Hashimoto and Sanmi Koyejo and Yejin Choi and Yu Sun and Xiaolong Wang},
      year={2025},
      eprint={2504.05298},
      archivePrefix={arXiv},
      primaryClass={cs.CV},
      url={https://arxiv.org/abs/2504.05298}, 
}

@misc{gs_dit_cvpr_25,
      title={GS-DiT: Advancing Video Generation with Pseudo 4D Gaussian Fields through Efficient Dense 3D Point Tracking}, 
      author={Weikang Bian and Zhaoyang Huang and Xiaoyu Shi and Yijin Li and Fu-Yun Wang and Hongsheng Li},
      year={2025},
      eprint={2501.02690},
      archivePrefix={arXiv},
      primaryClass={cs.CV},
      url={https://arxiv.org/abs/2501.02690}, 
}

@inproceedings{
fifodiffusion,
title={{FIFO}-Diffusion: Generating Infinite Videos from Text without Training},
author={Jihwan Kim and Junoh Kang and Jinyoung Choi and Bohyung Han},
booktitle={The Thirty-eighth Annual Conference on Neural Information Processing Systems},
year={2024},
url={https://openreview.net/forum?id=uikhNa4wam}
}

@inproceedings{
VAR,
title={Visual Autoregressive Modeling: Scalable Image Generation via Next-Scale Prediction},
author={Keyu Tian and Yi Jiang and Zehuan Yuan and BINGYUE PENG and Liwei Wang},
booktitle={The Thirty-eighth Annual Conference on Neural Information Processing Systems},
year={2024},
url={https://openreview.net/forum?id=gojL67CfS8}
}

@article{cogvideox,
  title={CogVideoX: Text-to-Video Diffusion Models with An Expert Transformer},
  author={Yang, Zhuoyi and Teng, Jiayan and Zheng, Wendi and Ding, Ming and Huang, Shiyu and Xu, Jiazheng and Yang, Yuanming and Hong, Wenyi and Zhang, Xiaohan and Feng, Guanyu and others},
  journal={arXiv preprint arXiv:2408.06072},
  year={2024}
}

@article{opensora,
  title={Open-sora: Democratizing efficient video production for all},
  author={Zheng, Zangwei and Peng, Xiangyu and Yang, Tianji and Shen, Chenhui and Li, Shenggui and Liu, Hongxin and Zhou, Yukun and Li, Tianyi and You, Yang},
  journal={arXiv preprint arXiv:2412.20404},
  year={2024}
}

@article{opensora2,
    title={Open-Sora 2.0: Training a Commercial-Level Video Generation Model in {\$}200k}, 
    author={Xiangyu Peng and Zangwei Zheng and Chenhui Shen and Tom Young and Xinying Guo and Binluo Wang and Hang Xu and Hongxin Liu and Mingyan Jiang and Wenjun Li and Yuhui Wang and Anbang Ye and Gang Ren and Qianran Ma and Wanying Liang and Xiang Lian and Xiwen Wu and Yuting Zhong and Zhuangyan Li and Chaoyu Gong and Guojun Lei and Leijun Cheng and Limin Zhang and Minghao Li and Ruijie Zhang and Silan Hu and Shijie Huang and Xiaokang Wang and Yuanheng Zhao and Yuqi Wang and Ziang Wei and Yang You},
    year={2025},
    journal={arXiv preprint arXiv:2503.09642},
}

@article{hunyuanvideo,
  title={Hunyuanvideo: A systematic framework for large video generative models},
  author={Kong, Weijie and Tian, Qi and Zhang, Zijian and Min, Rox and Dai, Zuozhuo and Zhou, Jin and Xiong, Jiangfeng and Li, Xin and Wu, Bo and Zhang, Jianwei and others},
  journal={arXiv preprint arXiv:2412.03603},
  year={2024}
}

@InProceedings{rollling-diffusion,
  title = 	 {Rolling Diffusion Models},
  author =       {Ruhe, David and Heek, Jonathan and Salimans, Tim and Hoogeboom, Emiel},
  booktitle = 	 {Proceedings of the 41st International Conference on Machine Learning},
  pages = 	 {42818--42835},
  year = 	 {2024},
  editor = 	 {Salakhutdinov, Ruslan and Kolter, Zico and Heller, Katherine and Weller, Adrian and Oliver, Nuria and Scarlett, Jonathan and Berkenkamp, Felix},
  volume = 	 {235},
  series = 	 {Proceedings of Machine Learning Research},
  month = 	 {21--27 Jul},
  publisher =    {PMLR},
  url = 	 {https://proceedings.mlr.press/v235/ruhe24a.html}
}

@article{diffusion_forcing,
  title={Diffusion forcing: Next-token prediction meets full-sequence diffusion},
  author={Chen, Boyuan and Mart{\'\i} Mons{\'o}, Diego and Du, Yilun and Simchowitz, Max and Tedrake, Russ and Sitzmann, Vincent},
  journal={Advances in Neural Information Processing Systems},
  volume={37},
  pages={24081--24125},
  year={2025}
}

@inproceedings{mncraft_dset,
  title={Temporally consistent transformers for video generation},
  author={Yan, Wilson and Hafner, Danijar and James, Stephen and Abbeel, Pieter},
  booktitle={International Conference on Machine Learning},
  pages={39062--39098},
  year={2023},
  organization={PMLR}
}

@article{
latte,
title={Latte: Latent Diffusion Transformer for Video Generation},
author={Xin Ma and Yaohui Wang and Xinyuan Chen and Gengyun Jia and Ziwei Liu and Yuan-Fang Li and Cunjian Chen and Yu Qiao},
journal={Transactions on Machine Learning Research},
issn={2835-8856},
year={2025},
url={https://openreview.net/forum?id=ntGPYNUF3t},
note={}
}
\bibliographystyle{tmlr}

\appendix

\clearpage
\appendix
\section{Appendix}

\subsection{Related Work}
\label{app_sec: rel_work}
\paragraph{Diffusion Models.} The superior generation quality of diffusion models has made them the de facto paradigm of choice for image and sequence generation \citep{stable_diff, videoldm}. Diffusion models, originated from score-based models \citep{score-1, score-2} and made popular with Denoising Diffusion Probabilistic Models (DDPMs) \citep{ddpm}, train a denoiser network that learns to reverse a corruption process that adds Gaussian noise to the data. The first diffusion pipelines used convolutional U-Nets \citep{unets} as the denoiser architecture. We employ the \textbf{Diffusion Transformer (DiT)} \citep{dit} architecture for our denoiser networks. DiT is the current SoTA on image generation. Instead of a U-Net denoiser, DiTs utilize a series of blocks with multi-headed self-attention. Another critical aspect of DiT is converting 2D images into a 1D sequence of tokens by patchifying the image, computing patch embeddings with positional encoding and ordering them in a sequence. This process allows the DiT to learn a strong self-attention prior over all spatial regions within the image. The prior forms the bedrock of our method as shown in later sections.

\paragraph{Video Generation.} Despite its importance, video generation lags behind image generation, primarily due to the high computational cost of processing large video tensors \citep{lvdm, videogpt, MoCoGAN, MoCOGANHD, PVDM, style-gan-v, digan, ddmi, alyosha_long_vid_gan, emu_video, animate_diff}. Most methods \citep{videogpt, MoCoGAN, MoCOGANHD, alyosha_long_vid_gan, videoldm, emu_video} model videos as large tensors, limiting maximum sequence length and incurring slow inference rates. Recent works using DiTs \citep{ditctrl_cvpr25, gs_dit_cvpr_25, ttt_vid_cvpr25} focus on architectural improvements for better conditioning; we consider these concurrent, but orthogonal to our goal of rethinking image sequence modeling. Approaches leveraging proxy models such as INRs \citep{style-gan-v, digan, ddmi} trade off efficiency for perceptual quality, yet still face limitations in scalability. Factorized generation has shown promise: Emu Video and AnimateDiff \citep{emu_video, animate_diff} split text-to-video into text-to-image and image-to-video stages, while LongVideoGAN \citep{alyosha_long_vid_gan} factorizes within a GAN-based framework but lacks support for arbitrary-length sequences or resolutions beyond $256{\times}256$.   

There exists a paucity of methods \citep{style-gan-v, lvdm, PVDM, tats} that attempt arbitrary length video generation. Of these, PVDM \citep{PVDM} and LVDM \citep{lvdm} are latent diffusion based approaches. Whereas TATS \citep{tats} uses a GAN and StyleGAN-V is a GAN approach paired with INRs. We compare with all these methods on the standard video generation task and with StyleGAN-V \citep{style-gan-v}, LVDM \citep{lvdm}, and TATS \citep{tats} on the arbitrary length video generation task to ensure that we cover the several different approaches taken to solve the problem. Ours is the first method to employ factorization in a self-attention powered diffusion regime for arbitrarily long image sequence generation to the best of our knowledge.   

\paragraph{Autoregressive sampling.} Autoregressive (AR) sampling entails employing information from previously generated samples to generate new samples. The rise of self-attention powered transformers \citep{txformer} has led to a wide array of AR generation applications \citep{txformer, gpt, xlnet} in Natural Language Processing (NLP). However, the technique remains under-utilized in the image sequence generation context, with only a handful of methods \citep{lvdm, magvitv2, nova, videogpt} making use of it. Of these, LVDM \citep{lvdm} is the most closely related to our work as it attempts to employ AR sampling for arbitrarily long video generation. Although, it does not make use of a grid-based formulation. We recognize NOVA \citep{nova} as a concurrent work and MAGVIT-v2 \citep{magvitv2} employs Large Language Models (LLMs) for video generation which is an orthogonal research direction. Therefore, we omit NOVA \citep{nova} and MAGVIT-v2 \citep{magvitv2} in our comparative studies. 

\paragraph{Image Sequence Generation.} We address image sequence generation beyond conventional video tasks. For instance, lung CT \citep{lung_ct, ddgsct_ct, diffusionblend_ct, miccai_3d_ct} and brain MRI \citep{brain_mri} in 3D medical imaging, where modeling inter-slice dependencies is critical. Regarding 3D CT volume synthesis, GenerateCT \citep{GenCT} is the only method reporting spatiotemporal consistency on public 3D CT data, making it our primary baseline. Notably, GenerateCT employs a complex, text-conditional three-stage pipeline. Whereas our approach is simple and unconditional.      

\subsubsection{Concurrent work and Very Large-scale Models.}  
\label{app_sec:concurr_work}
We treat very large-scale models ($\geq$ 2B parameters) such as Open-Sora \citep{opensora, opensora2}, Hunyan Video \citep{hunyuanvideo}, CogVideoX \citep{cogvideox}, FIFO-diffusion (all variants implemented by the authors) \citep{fifodiffusion}, \citep{cogvideox}, and VAR \citep{VAR} beyond the scope of comparison with this work. We also exclude comparisons with models trained on combinations of multiple datasets or those trained on a single very large (more than a few hundred thousand datapoints) datasets. We make this choice for two reasons. First, it is computationally intractable to work with them within our compute budget. Second, the comparison is unfair for our model. As far as concurrent work is concerned, LATTE \citep{latte} stands out among the plethora of related works. Although it is similar to our approach in using the DiT \citep{dit}, it's objective of quality maximization of fixed-length videos is fundamentally different to ours. Our approach is more about devising a data modeling scheme that supports arbitrary-length, efficient, and generalizable image-sequence generation than about fixed-length video quality maximization.

\subsection{Preliminary on Denoising Diffusion Models}
\label{app_sec: prelim}
We employ Denoising Diffusion Probabilistic Models (DDPMs) ~\citep{ddpm} to learn and sample from the target distributions. DDPMs generate samples by learning to invert the process of information corruption by adding Gaussian noise. The forward diffusion corrupts the data, which the learned model reverses to synthesize new samples. The forward process is characterized by: $x_t=\sqrt{\bar{\alpha}_t} x_0+\sqrt{1-\bar{\alpha}_t} \epsilon_t, \text { where } \epsilon_t \sim \mathcal{N}(0, \mathbf{I})$ and the reverse process is characterized by: $\mathbf{x}_{t-1}=\frac{1}{\sqrt{\alpha_t}}\left(\mathbf{x}_t-\frac{1-\alpha_t}{\sqrt{1-\overline{\alpha_t}}} \boldsymbol{\epsilon}_\theta\left(\mathbf{x}_t, t\right)\right)+\sigma_t \mathbf{z}$, where $z \sim \mathcal{N}(0, \mathbf{I})$ and $\boldsymbol{\epsilon}_\theta$ is the noise predicted by the learned model parametrized by $\theta$.

\subsection{Model Architectures}
\label{app_sec: model_arch}
Our Stage 1 (\emoticon{images/one.png}) model is a DiT \citep{dit} wherein class conditioning is removed, and only timestep conditioning is retained when used as a denoiser in our DDPM \citep{ddpm} training process. The hyperparameters employed in training Stage 1 are listed below:
\begin{itemize}
    \item DiT variant: DiT-XL/2
    \item Training resolution: 512
    \item Model depth: 28
    \item Embedding dimension: 1152
    \item Patch size: 2
    \item Number of self-attention heads: 16
\end{itemize} 

Whereas Stage 2 (\emoticon{images/2.png}) is a vanilla DiT with appropriate modifications to use low-res (degraded) images as class conditioning. The subtle modifications are outlined below. Stage 2 shares the same hyperparameters as stage 1.
\subsubsection{Specifics of our Stage 2 architecture for Super-resolution}
\label{app_sec: SR}
As illustrated in Figure \ref{fig: training-method} (b), we super-resolve the low-resolution frames using a conditional DiT model. In this case, we use the DiT model with certain modifications since we are generating a single, high-resolution (HR) image, conditioned on its low-resolution (LR) counterpart. More specifically, we obtain LR images from our training datasets of HR images by performing a combination of degradations (noise addition and 
\begin{figure}[ht]
    \centering
    \includegraphics[scale=0.33]{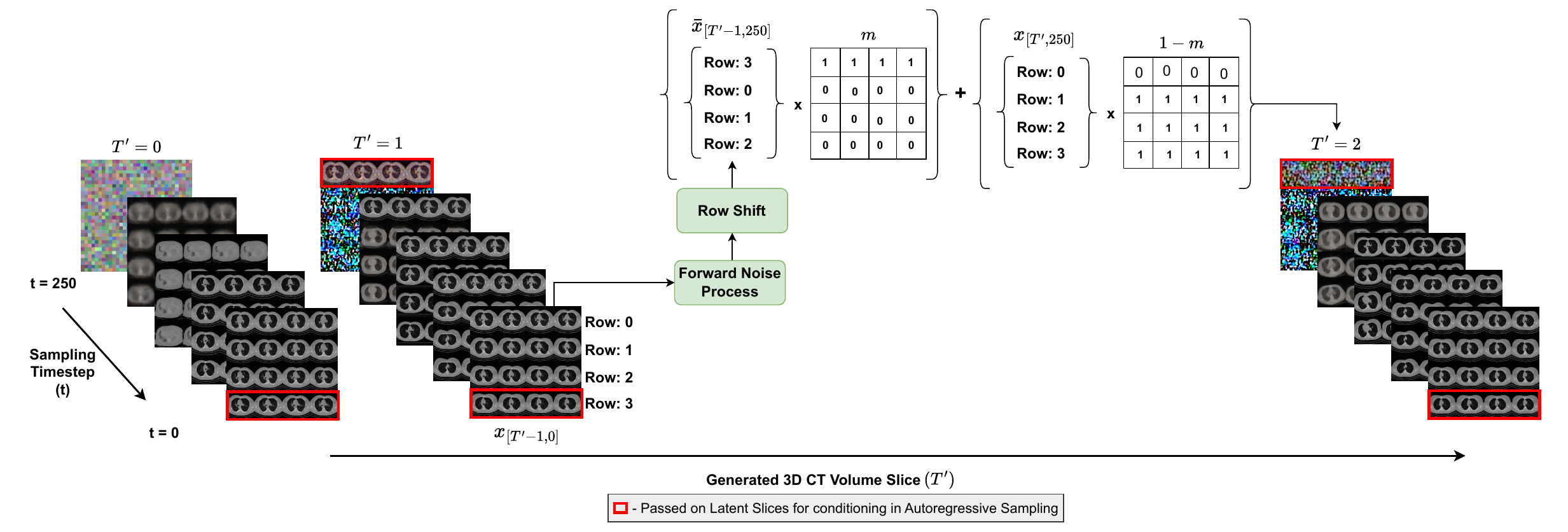}
    \caption{We illustrate the \texttt{Row Shift and Masking} operations used to form the \textit{control signal} in our Grid-based Autoregressive sampling algorithm to transfer the last row from the previous iteration as the first in the next one while sampling grid images corresponding to subsampled synthetic 3D CT Volumes similar to ones from the CT-RATE dataset. The one-row control signal is only shown for brevity and clarity. In practice, we use three-row control signals in step 1 sampling for all our experiments.}
    \label{fig: recur-alter}
\end{figure}
blurring) on the HR images. Our training dataset for the task now comprises several $\{$LR, HR$\}$ pairs. For each pair, the SR model's goal is to learn to generate HR images conditioned on the corresponding LR input. We train it to do so by embedding both LR and HR images in the VAE's \citep{vae} latent space, concatenating the two latents,  projecting the concatenated latent on the original hidden dimension, and training the DiT to generate the embedding for HR given the obtained projected embedding as input. We apply conditioning to the DiT via adaptive layer norm. We summarize the process of going from a coarse synthetic grid image comprising subsampled frames to highly photorealistic and motion-preserving individual frames in Figure \ref{fig: inference}. 

\subsection{Grid-based Frame Modeling - a formal perspective}
\label{app_sec: grid-frame-modeling}
Here, we elaborate upon the process of obtaining grid images from image sequence data. These grid images are later used in the low-res generation aspects of our work. As illustrated in Figure \ref{fig: training-method}, given an $  \bar{N}\times  H\times W$ tensor data point, we first down-sample it to $  \bar{N}\times H/K \times W/K$ dimensions using bicubic interpolation. Subsequently, we extract $K^{2}$-length sub-sequences along the tensor's channel dimensions such that the $n^{th}$ sub-sequence comprises indices $\{n,n+1,n+2\dots n+K^{2}\}$ along the channel dimension. Finally, for grayscale images, all elements of the sub-sequence are repeated three times along the channel dimension and concatenated, while preserving their ordering, to form a grid tensor with dimensions $3 \times H \times W$. Therein, the grid elements $(0,0)$, $(0,1)$, and $(K-1,K-1)$  denote the CT slices indexed by $n$, $n+1$, $n+K^{2}$ along  the channel dimension of the sub-sequence, respectively.

\subsection{Additional details on Grid-based Autoregressive sampling}
\label{app_sec:grid-based-ar-sampling}
\paragraph{Notation pertaining to our sampling scheme.} $
\tX_{[T',t]}, \tX_{[T',t]}, \tX_{prev}, \tX_{next}, \bar{X}, m, m_{prev}, m_{current}, \text{and}\,\, \\ m_{next}$ denote a sample at iteration $T'$ and diffusion timestep $t$ in step 1, a sample at iteration $T''$ and diffusion timestep $t$ in step 2, the previous sample from step 1 at iteration $T''$ of step 2, the next sample from step 1 at iteration $T''$ of step 2, forward noised version of a sample $X$, a binary mask representing a grid with $K=4$ having first three rows set to 1 and last one set to 0, a similar binary mask having first row set to 1 and rest set to 0, a similar binary mask having rows 1,4 set to 0 and rows 2,3 set to 1, a similar binary mask having last row set to 1 and rest set to 0.   

\paragraph{A Formal Perspective on Our Sampling Scheme.} As summarized by Algorithms \ref{algo:grid-ar-sampling-step-1} and \ref{algo:grid-ar-sampling-step-2}, the process of generating the grid-image $T'$ in Step 1 starts with generating the first grid image $\tX_{[0,0]}$ per the vanilla DiT sampling procedure. This is followed by several autoregressive sampling iterations until the grid image $T'$ is arrived at. Each reverse diffusion timestep $t$ of every autoregressive iteration $T'$ entails:
\begin{algorithm}[H]
\caption{Grid-based Autoregressive Sampling Step 1 (Coarse Generation)}
\label{algo:grid-ar-sampling-step-1}
\textbf{Input:} $\tX_{[0,0]} \,\,\,\,\triangleright$ The first grid image generated via standard DDPM sampling for $\mathcal{T}_s$ steps.\\
    \textbf{Output:} $\tV' \,\,\,\,\triangleright$ A coarsely-coherent sequence of grid images starting with $\tX_{[0,0]}$.\\
\begin{algorithmic}[1] 
\State $\tV' \leftarrow \{\}$
        \For{$T' = 1, 2, \dots, N$}
            \State $\tX_{[T', \mathcal{T}]} \sim \mathcal{N}(0, I)$
            \For{$t = \mathcal{T}_s, \mathcal{T}_s-1, \mathcal{T}_s-2, \dotsc, 1$}
                \If{$t > 1$}
                    \State $\epsilon \sim \mathcal{N}(0, \mI)$
                \Else
                    \State $\epsilon \leftarrow 0$
                \EndIf
                \State $\bar{\tX}_{[T'-1, t-1]} \leftarrow 
                    \sqrt{\bar{\alpha}_t}\,\tX_{[T'-1, 0]} 
                    + \sqrt{1-\bar{\alpha}_t}\,\epsilon$
                \State $\bar{\tX}_{[T'-1, t-1]} \leftarrow 
                    \text{RowShift}(\bar{\tX}_{[T'-1, t-1]})$
                \State $\tX_{[T', t-1]} \leftarrow 
                    \sigma_t\,\epsilon 
                    + \frac{1}{\sqrt{\alpha_t}}\Big(
                        \tX_{[T', t]} 
                        - \frac{\beta_t}{\sqrt{1-\bar{\alpha}_t}}\,
                          \epsilon_{\theta_{1}^{\ast}}(\tX_{[T', t]}, t)
                      \Big)$
                \State $\tX_{[T', t-1]} \leftarrow 
                    (m \odot \bar{\tX}_{[T'-1, t-1]})
                    + ((1-m)\odot \tX_{[T', t-1]})$
            \EndFor
            \State $\tV' \leftarrow \tV' \cup \{\tX_{[T',0]}\}$
        \EndFor
        \State \Return $\tV'$
\end{algorithmic}
\end{algorithm}

\begin{algorithm}[H]
\caption{Grid-based Autoregressive Sampling Step 2 (Interpolation for Temporal Super-resolution)}
\label{algo:grid-ar-sampling-step-2}
\textbf{Input:} $\tV' = \{\tX_{[T',0]}: T' \in [0, N]\} \,\,\,\,\triangleright$ The output sequence from Step 1.\\
    \textbf{Output:} $\tV'' \,\,\,\,\triangleright$ A spatially coarse sequence of grid images with superior temporal resolution than Step 1.\\
\begin{algorithmic}[1] 
\State $\tV'' \leftarrow \{\}$
        \For{$T'' = 0, 1, \dots, N-1$}
            \State $\tX_{[T'', \mathcal{T}]} \sim \mathcal{N}(0, I)$
            \State $T' = T ''$
            \State $\tX_{prev} = \tX_{[T',0]}, \tX_{next} = \tX_{[T'+1,0]}$
            \For{$t = \mathcal{T}_s, \mathcal{T}_s-1, \mathcal{T}_s-2, \dotsc, 1$}
                \If{$t > 1$}
                    \State $\epsilon \sim \mathcal{N}(0, \mI)$
                \Else
                    \State $\epsilon \leftarrow 0$
                \EndIf
                \State $\bar{\tX}_{prev} \leftarrow 
                    \sqrt{\bar{\alpha}_t}\,\tX_{prev} 
                    + \sqrt{1-\bar{\alpha}_t}\,\epsilon$
                \State $\bar{\tX}_{next} \leftarrow 
                    \sqrt{\bar{\alpha}_t}\,\tX_{next} 
                    + \sqrt{1-\bar{\alpha}_t}\,\epsilon$    
                \State $\tX_{[T'', t-1]} \leftarrow 
                    \sigma_t\,\epsilon 
                    + \frac{1}{\sqrt{\alpha_t}}\Big(
                        \tX_{[T'', t]} 
                        - \frac{\beta_t}{\sqrt{1-\bar{\alpha}_t}}\,
                          \epsilon_{\theta_{1}^{\ast}}(\tX_{[T'', t]}, t)
                      \Big)$
                \State $\tX_{[T'', t-1]} \leftarrow 
                    (m_{prev} \odot \text{RowShift}(\bar{\tX}_{prev}))
                    + (m_{current}\odot \tX_{[T'', t-1]}) + (m_{next} \odot \bar{\tX}_{next})$
            \EndFor
            \State $\tV'' \leftarrow \tV'' \cup \{\tX_{[T'',0]}\}$
        \EndFor
        \State $\tV'' \leftarrow \tV' \cup \tV''$ $\,\,\,\,\triangleright$ Frames obtained by splitting the grid images in $\tV'$ and $\tV''$ and retaining unique elements only are combined by inserting newly interpolated frames between previously generated frames.
        \State \Return $\tV''$
\end{algorithmic}
\end{algorithm}
(1) adding appropriate noise to the previous generated grid image $\tX_{[T'-1,0]}$ per the forward process, (2) obtaining its row-shifted and masked version $\bar{\tX}_{[T'-1, t-1]}$, (3) combining that version with a tensor obtained 
\begin{figure}[H]
    \centering
    \includegraphics[width=\linewidth]{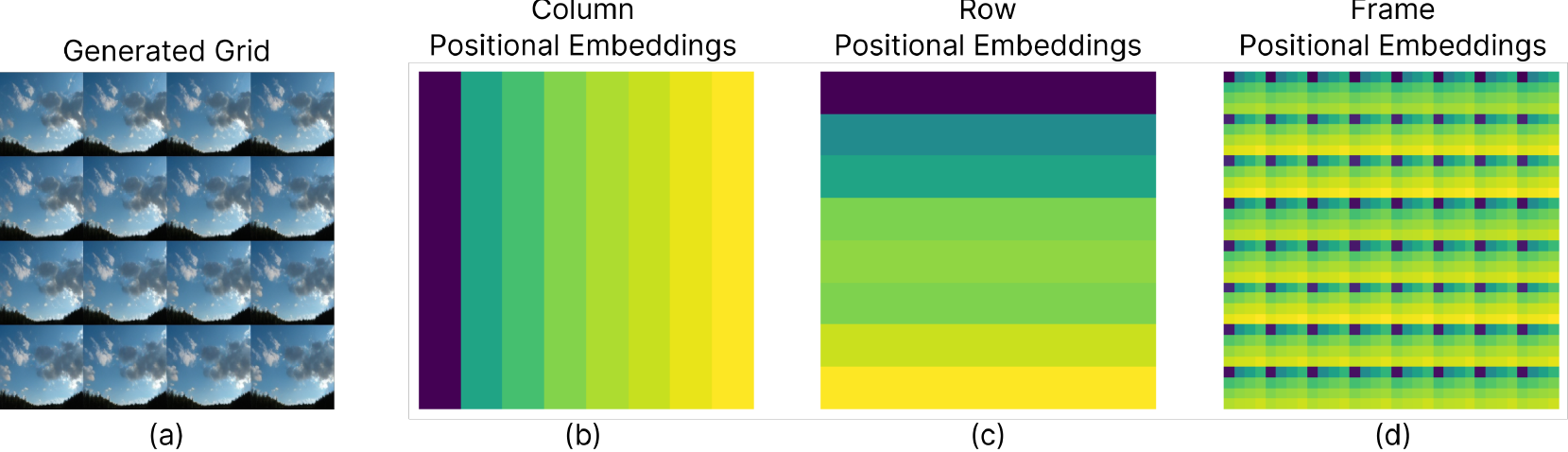}
    \caption{(a) In image grid generated from our SkyTimelapse model. In (b), (c), and (d), we contrast how the different positional embeddings encode information about the (x,y) location of each patch as well as its temporal location within the sequence. Here, (b) and (c) represent different 2D positional embeddings, whereas (d) represents 3D positional embeddings. We make use of the (d) 3D (frame) positional embeddings in our formulation to make sure the correspondence between grid elements is captured perfectly. Similarities between (a) and (d) are indicative of the suitability of 3D positional embeddings to capture sequential information structured as grids.}
    \label{fig: pos-emb}
\end{figure}
\begin{algorithm}[H]
\caption{Diffusion-driven 3D CT Volume Denoising}
\label{algo: 3d-ct-volume-denoising}
\textbf{Input}: $\tX_{[n,0]}$ (Nosiy Image Latents)\\
\textbf{Output}: $\hat{\tX}_{0}$ (The Denoised Latents)\\
\begin{algorithmic}[1] 
\For{$t= \mathcal{T}, \dotsc, 1$}
    \State $\epsilon \sim \mathcal{N}(0, I)$ if $t > 1$, else $\epsilon = 0$
    \State $\tX_{[n, t]} = \sqrt{\bar{\alpha}_{t+1}} \tX_{[n, 0]} + \sqrt{(1-\bar{\alpha}_{t+1})} \epsilon$
    \State $\hat{\tX}_{t-1} =
          \frac{1}{\sqrt{\alpha_t}}\left(\tX_{[n, t]} - \frac{\beta_t}{\sqrt{1-\bar\alpha_t}} \epsilon_{\theta_{1}^{\ast}}(\tX_{[n, t]}, t) \right)
          + \sigma_t \epsilon$
\EndFor
\State \textbf{return} $\hat{\tX}_{0}$
\end{algorithmic}
\end{algorithm}
by denoising $\tX_{[T', \mathcal{T}]}\sim\mathcal{N}(0,\mI)$ for $t-1$ timesteps. The process continues for $\forall \,t\in[0,\mathcal{T}]$ and $\forall\, T'\in[1,N]$. In effect, the binary mask $m$ acts as a gating mechanism that decides the amount of previously generated information used at a particular autoregressive sampling iteration. All operations are performed in the DiT latent space, with scale factors applied for reduced dimensions. The latent encoding and decoding steps are omitted for brevity. 

The core operations performed in step 2 are similar to step 1 except for the fact that they now result in interpolated frames due to the nature of ordering in our grid-based formulation and the model's priors. 

The binary masks $m, m_{prev}, m_{current}, \text{and} \,\,\, m_{next}$ are a $K \times K$ matrices scaled appropriately to latent space dimensions with all elements as described in the Notation pertaining to our sampling scheme. This is further elucidated by Figure \ref{fig: recur-alter} which also illustrates the \texttt{RowShift} operation. We set $\mathcal{T_{s}}, H, W,$ and $K$ to be 250, 512, 512, and 4 in all our experiments on the CT-RATE dataset \citep{ct-rate-1, ct-rate-2}. Whereas for experiments on the SkyTimelapse \citep{skydset} and Taichi \citep{taichi_dset} dataset, we change $H$ and $W$ to 256.
\subsection{3D positional embeddings}
\label{app_sec: 3d_pos_embed}
In Table~\ref{tab:pos-embed-ablation}, we rerun our experiments without the 3D positional embeddings that we propose adding to the image grids. Without encoding the position within the sequence, the generated patches exhibit worse temporal consistency, which is evident by the reduction in FVD observed. Further, in Figure~\ref{fig: pos-emb} we visualize how the 3D positional embeddings we use encode both the spatial location and location within the sequence for each individual image patch the DiT operates on. Thereby playing a crucial part in our Grid-based formulation.

Formally, our 3D positional embeddings use a broad mathematical formulation similar to the one employed by fixed  $sin-cos$ positional embeddings from \citet{txformer}. Our embeddings differ only in that we embed each latent dimension separately, followed by stacking them together along the positional embedding dimension. The equations below summarize our embedding scheme:
\begin{align}
\operatorname{PosEmbed}(p_{d_l}, p_h, p_w) 
&= \left[
    \mathbf{e}^{(d_l)}(p_d);\,
    \mathbf{e}^{(h)}(p_h);\,
    \mathbf{e}^{(w)}(p_w)
\right] \in \mathbb{R}^{D} \label{eq:posembed} \\[0.8em]
\mathbf{e}(p) 
&= \left[\,
    \sin(p \cdot \omega_0),\, \ldots,\, \sin\!\left(p \cdot \omega_{\frac{d}{2}-1}\right),\,
    \cos(p \cdot \omega_0),\, \ldots,\, \cos\!\left(p \cdot \omega_{\frac{d}{2}-1}\right)
\,\right] \label{eq:embedfunc} \\[0.8em]
d_{d_l}=d_h=d_w &= \left\lfloor \frac{D}{3} \right\rfloor \label{eq:dh}
\end{align}
Wherein $\operatorname{PosEmbed}(p_d, p_h, p_w)$ represents the overall positional embedding at a particular embedding dimension index ($d_l$), latent height index ($h$), and latent width index ($w$) for a total embedding length $D$. Eq. \ref{eq:embedfunc} further elaborates upon each individual 1D embedding component in Eq. \ref{eq:posembed}. Each individual-dimensional embedding bears a formulation similar to \citet{txformer} with $p$ being an indexing variable for positions. Whereas Eq. \ref{eq:dh} clarifies that each embedding individual 1D embedding dimension is allotted an equal length.

\subsection{Experimental Setup}
\label{app_sec: exp_setup}
We carefully curate our experiments with a threefold objective: evaluate GriDiT's performance for high-quality, arbitrary-length image sequence synthesis across diverse data regimes; justify our design choices; and elucidate the mechanisms underlying our method. Here, we provide comprehensive information about the various settings and design choices employed during these experiments.

\subsubsection{Dataset Details}
\label{app_sec: dset}
We conduct our evaluations on three significantly different datasets that are widely used in baselines established by the prior art. (1) \textbf{The SkyTimelapse dataset} \citep{skydset}, which comprises timelapse videos of skies in different lighting scenarios and associated ground imagery. Performance on this dataset is primarily indicative of generating sequences with long-range temporal consistency and photorealism of frames, as the dataset does not contain rapid motion or occlusions. We work with this dataset at $256\times256$ resolution and use the provided train and test splits. (2) \textbf{The CT-RATE dataset} \citep{ct-rate-1, ct-rate-2} that contains several 3D CT Volumes collected from real patients. These Volumes are sequences wherein the axis of variation is non-temporal. Performing well on this dataset requires accurate modeling of high-frequency structure (anatomy) and textures, as well as long-range consistency. We learn to generate sequences from this dataset at a resolution of $512\times512$. We use exactly the same preprocessing, train split, and test split as GenerateCT \citep{GenCT}. (3) \textbf{The Taichi dataset} \citep{taichi_dset} that was originally proposed for human action recognition and requires accurate modeling of motion at 
both coarse and fine scales to produce high-quality results. We use the dataset in its standard $256\times256$ resolution using its train and test splits.
\subsubsection{Training} 
\label{app_sec: trg}
We train two DiT (DiT-XL/2) \citep{dit} models corresponding to our Stage 1 (\emoticon{images/one.png}) and Stage 2 (\emoticon{images/2.png}) models, respectively, for each dataset as described in section \ref{sec: core_method}. Both stages of all our models were finetuned starting from the DiT \citep{dit} weights pretrained on ImageNet \citep{imagenet}, with a linear warmup schedule taking the learning rate from $10^{-6}$ to $10^{-4}$ over $10^{4}$ warmup 
\begin{table}[h]
\centering
\caption{Denoising 3D CT Volumes. We compare different denoising methods under varying noise levels. Our method outperforms the baselines in all settings, demonstrating superior denoising capability. ($\mathcal{N}(0,\sigma^{2})$: Noise (degradation) process with mean $=0$ and variance $=\sigma^{2}, $\textuparrow : higher is better.)}
\label{tab:denoising-comparison}
{\small
\begin{tabular}{l
                cc
                cc}
\toprule
\textbf{Denoising Method} 
& \multicolumn{2}{c}{\textbf{$\mathcal{N}(0, 25)$}} 
& \multicolumn{2}{c}{\textbf{$\mathcal{N}(0, 100)$}} \\
\cmidrule(lr){2-3} \cmidrule(lr){4-5}
& PSNR (dB)$\uparrow$ & SSIM$\uparrow$
& PSNR (dB)$\uparrow$ & SSIM$\uparrow$ \\
\midrule
Bilateral Filtering 
& 16.5  & 0.308
& 16.12 & 0.294 \\
GenerateCT 
& 23.81 & 0.357
& 23.41 & 0.350 \\
\rowcolor{Gray}
\textbf{Ours} 
& \textbf{41.26}  & \textbf{0.855}
& \textbf{34.65} & \textbf{0.758} \\
\bottomrule
\end{tabular}
}
\end{table}

\begin{table}[h]
\centering
\caption{We contrast our SR module's performance with that of the SoTA (SinSR) at different SR scales. Our model outperforms the SoTA in all settings, signifying its efficacy. (\textuparrow : higher is better. \textdownarrow : lower is better. $^{\dagger}$ : fine-tuned model metrics. $^{\ast}$ : vanilla model metrics. \textbf{Bold}: best entry.}
\label{tab:sr-comparison}
\resizebox{\textwidth}{!}{%
{\small
\begin{tabular}{l
                cc
                cc
                cc
                cc}
\toprule
\textbf{Method} 
& \multicolumn{2}{c}{\textbf{SkyTimelapse ($\times$2)}} 
& \multicolumn{2}{c}{\textbf{CT-RATE ($\times$4)}} 
& \multicolumn{2}{c}{\textbf{Taichi ($\times$2)}} 
& \multicolumn{2}{c}{\textbf{Taichi ($\times$4)}} \\
\cmidrule(lr){2-3} \cmidrule(lr){4-5} \cmidrule(lr){6-7} \cmidrule(lr){8-9}
& PSNR (dB)$\uparrow$ & FVD-16$\downarrow$
& PSNR (dB)$\uparrow$ & FVD-16$\downarrow$
& PSNR (dB)$\uparrow$ & FVD-16$\downarrow$
& PSNR (dB)$\uparrow$ & FVD-16$\downarrow$ \\
\midrule
SinSR 
& 29.8$^{\ast}$  & 71.7$^{\ast}$
& 21.42$^{\dagger}$ & 593.23$^{\dagger}$
& 34.52$^{\ast}$ & 138.67$^{\ast}$
& 31.93$^{\ast}$ & 119.48$^{\ast}$ \\
\rowcolor{Gray}
\textbf{Ours} 
& \textbf{31.982}  & \textbf{64.08}
& \textbf{29.48}   & \textbf{383.45}
& \textbf{35.28}   & \textbf{134.654}
& \textbf{33.48}   & \textbf{118.919} \\
\bottomrule
\end{tabular}
}
}
\end{table}

\citep{warmup} iterations. Self-attention based model architectures have been shown \citep{big_batchsz} to benefit from larger batch sizes. Therefore, we made use of engineering methods such as 
\begin{wrapfigure}{r}{0.5\textwidth}
    \centering
    \includegraphics[width=0.48\textwidth]{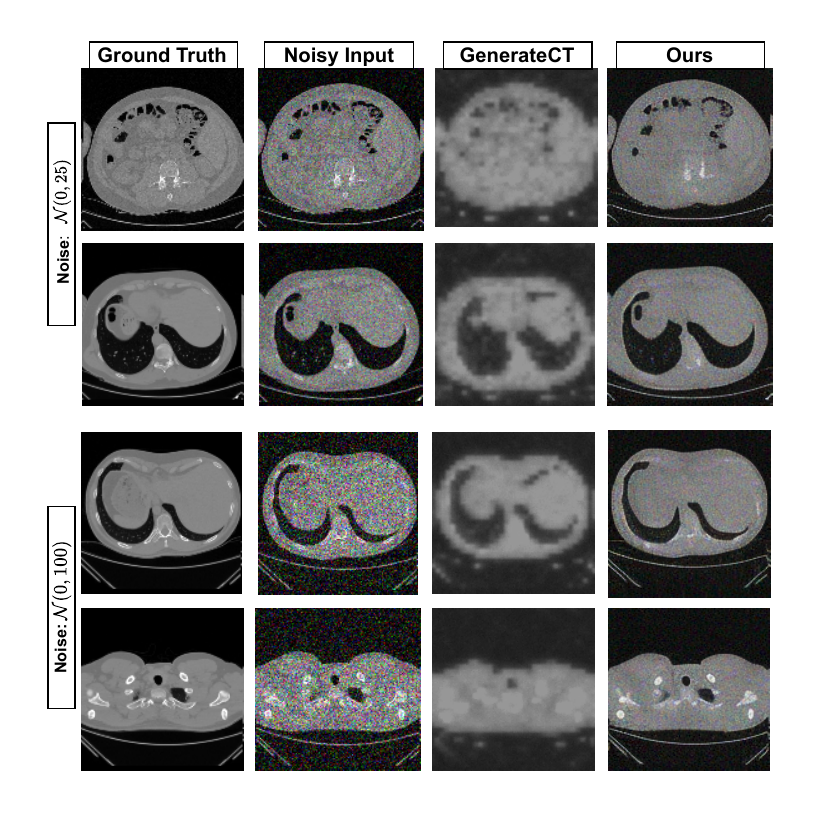}
    \caption{Employing our denoising diffusion pipeline for denoising 3D CT Volumes yields substantially superior performance than the baseline \citep{GenCT}. Thereby, suggesting a significantly stronger modeling ability, which enables our method to perform a task that is yet to emerge in the literature.}
    \label{fig: denoising-qual}
    \vspace{-10mm}
\end{wrapfigure}
mixed-precision training \citep{fp-16}  and gradient accumulation \citep{grad_acc} in our mini-batch optimization to train at larger batch sizes than those permitted by hardware constraints. We borrowed other elements of the training recipe viz. the objective function and the optimizer from DiT \citep{dit} to train our models. We used a single NVIDIA RTX A6000 GPU for training all models. 

\subsubsection{Sampling and Evaluation}
\label{app_sec: sampling_eval}
We use our Grid-based Autoregressive sampling algorithm paired with the SR method with different control signal settings for generating sequences in all our experiments. We evaluate our method on the quality of the synthesized sequences as well as their consistency. We chose the widely accepted FVD \citep{fvd} and FID \citep{fid} metrics for our experiments. We use the I$_3$D \citep{i3d} feature extraction backbone for computing FVD. We employ the evaluation pipeline established by StyleGAN-V \citep{style-gan-v} for computing FVD and FID. We measure the inference time in seconds (s) on an NVIDIA A100 GPU as a metric for efficiency. We do not provide metrics for intractable methods, viz., methods that do not support the experiment in question, do not provide their implementation, or yield severely poor qualitative performance when implemented. 

Finally, we bring forth two critical points that are important with respect to evaluation. First, we use unconditional Stage 1 sampling in all experiments in the paper. Second, in all experiments where the length of synthetic videos is greater than that of videos in the test set, we looped the test set videos cyclically to have them all reach the experimental length and computed metrics with respect to them. 
\begin{table}[ht]
\centering
\caption{Our model outperforms the SoTA on video generation on 16 and 128 length sequences from the SkyTimelapse dataset. Our model is significantly better in terms of perceptual quality and spatiotemporal consistency (FVD) and $>2.5\times$ faster. ($\downarrow$ : lower is better. $^\dagger$ and $^\ast :$  numbers reported by StyleGAN-V \cite{style-gan-v} and DDMI \cite{ddmi}, respectively. \textbf{Bold}: best entry. underline : second best entry.)}
\label{tab:main-quant-sky}
{\small
\begin{tabular}{lcccc}
\toprule
\multirow{3}{*}{\textbf{Method}} & \multicolumn{2}{c}{\textbf{16 Frames}}  & \multicolumn{2}{c}{\textbf{128 Frames}} \\ 
\cmidrule{2-5}
& \textbf{FVD}\textdownarrow & \textbf{Sampling Time (s)} \textdownarrow & \textbf{FVD}\textdownarrow & \textbf{Sampling Time (s)} \textdownarrow \\
\midrule
VideoGPT   & 222.7$^{\dagger}$ & 58.56 & -     & -      \\
MoCoGAN    & 206.6$^{\dagger}$ & -     & 575.9$^{\dagger}$ & -      \\
MoCoGAN-HD & 164.1$^{\dagger}$ & 77.8  & 878.1$^{\dagger}$ & -      \\
LVDM       & 95.2$^{\ast}$  & 91.75 & 233.4 & 273.4 \\
PVDM       & 71.46$^{\ast}$ & \underline{47.6}  & -     & -      \\
DIGAN      & 83.11$^{\dagger}$ & -     & \underline{196.7$^{\dagger}$} & -      \\
StyleGAN-V & 79.52$^{\dagger}$ & 62    & 197$^{\dagger}$   & \underline{243.25} \\
DDMI       & \underline{66.25$^{\ast}$} & -     & -     & -      \\
\rowcolor{Gray}
\textbf{Ours }      & \textbf{64.078} & \textbf{4.8} & \textbf{183.4} & \textbf{92.55} \\
\bottomrule
\end{tabular}}
\end{table}
\subsection{3D CT Volume Denoising}
\label{app_sec: denoising}
The fact that our method learns a strong self-attention prior to the denoising diffusion process and can work with arbitrary-length sequences prompted us to attempt image sequence (3D CT) Volume denoising. 3D CT volume denoising is a particularly relevant problem for two reasons. First, 3D CT Volumes are frequently corrupted by noise that creeps in due to improper calibration of CT scan machines or improper preprocessing as mandated by the instrument type. Second, to the best of our knowledge, this problem has not been addressed by any other denoising diffusion-based method. Given a noisy CT volume, we start by converting a series of subsequences of the noisy volume to grid-images encoded into our latent space employed for diffusion, per our grid-based modeling procedure. We denote these latents as $x_{[n,0]}$. Subsequently, we denoise the latents corresponding to the noisy input signal per the procedure summarized by Algorithm \ref{algo: 3d-ct-volume-denoising}. Effectively, we apply appropriate forward diffusion steps to the noisy signal and then perform reverse diffusion steps via our learned stage 1 model. Finally, we perform SR on the split grid elements using our learned Stage 2 and collate the slices together to get the denoised outputs. We prepare baselines for comparison using GenerateCT and standard bilateral filtering. GenerateCT does not support denoising. So, we used its diffusion-based Super-resolution block to perform denoising without text prompts to form a baseline. Our method outperforms GenerateCT and standard bilateral filtering in terms of denoised volume quality as presented in Table \ref{tab:denoising-comparison} and Figure \ref{fig: denoising-qual}, although it does lose out on a few high-frequency details in the ground truth image. These results are important as they further attest to our modeling and learning paradigm's efficacy in effectively representing image sequences.
\subsection{Custom Super-resolution} 
\label{app_sec: custom_sr_justification}
We present a quantitative comparison of our model's stage (\emoticon{images/2.png}) with a SoTA diffusion-based SR method, SinSR \citep{sinsr}, in Table \ref{tab:sr-comparison}. We finetuned SinSR for comparison on the CT-RATE dataset since CT Volume slices are reasonably out of distribution with respect to its training data. We used its pre-trained variant for other experiments. Our methods' superiority over the SoTA underscores its utility in making our method work. Furthermore, upon qualitative observation, we found that it is more suitable than the SoTA in retaining high-frequency details while performing SR. This attribute is hugely significant in critical domains such as 3D CT Volume imaging. Thereby, establishing that generative SR furthers our proposed method's applicability to general image sequence synthesis.
\subsubsection{On the degradation scheme employed to train our Stage 2 model \texorpdfstring{(\emoticon{images/2.png})}{ }}
\begin{figure}
    \centering
\includegraphics[width=\textwidth]{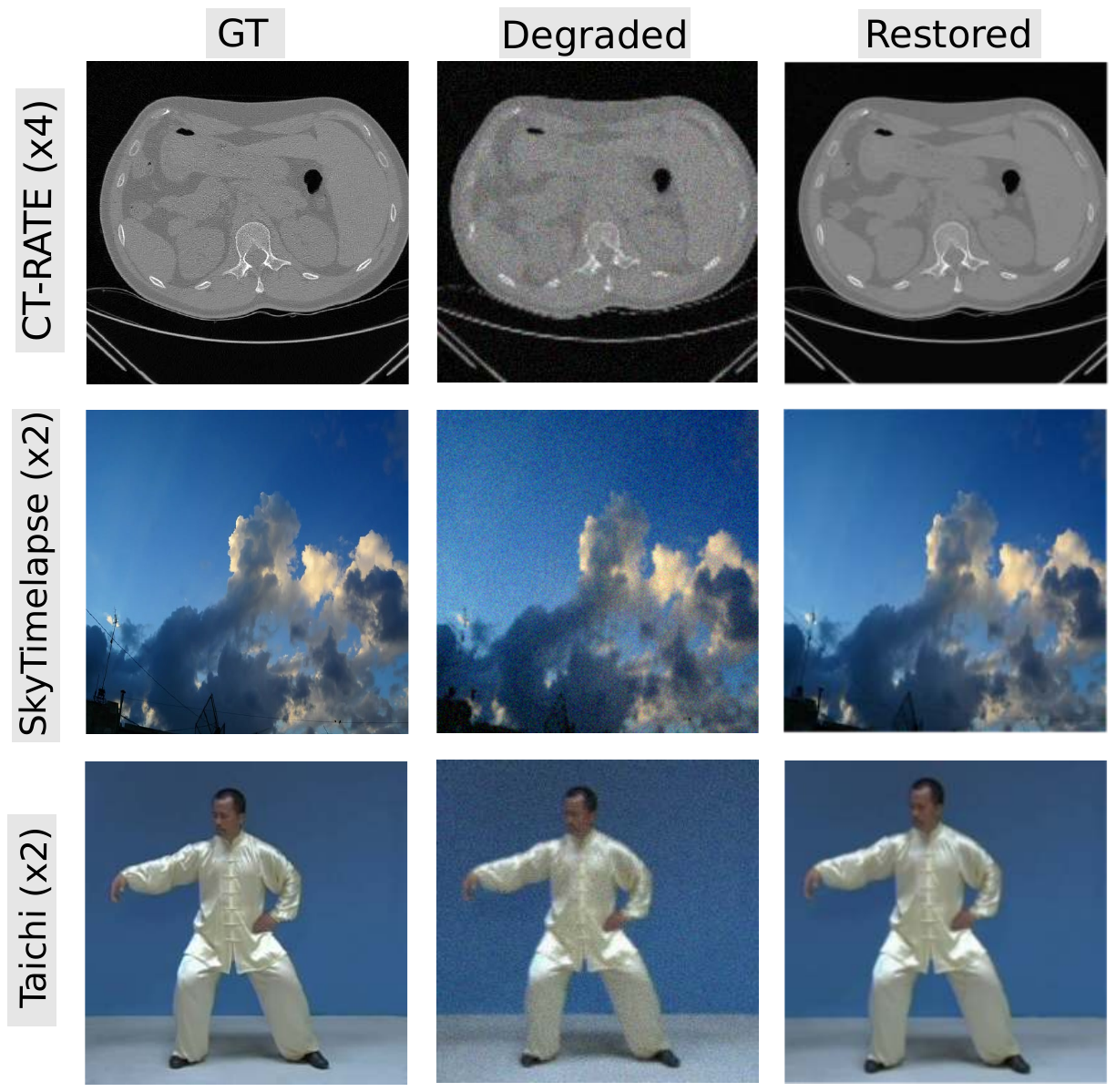}
    \caption{An overview of the surrogate restoration task performed by our method's stage 2 (\emoticon{images/2.png}) to perform individual frame refinement. (GT: Ground Truth).}
    \label{fig: sr-surrogate}
\end{figure}

As described in section 3.4 of the paper, our stage 2 model learns to refine individual frames by performing the surrogate task of restoring appropriately degraded frames. To that end, ground truth frames are first downsampled and then upsampled by the required scaling factor to simulate degradation caused by lossy super-resolution. We also add variable amounts of Gaussian noise to each image in the training dataset to account for losses caused by learning to model motion at low resolution. Consequently, these `degraded' frames are used as conditioning signals to generate their corresponding `restored' frames or the original ground truth frames. We use bicubic interpolation in all our scaling operations. At inference, refining a low-resolution frame entails upsampling it via bicubic interpolation and then using it to condition the stage 2 model's generation that yields the corresponding high-resolution frame. Figure \ref{fig: sr-surrogate} illustrates the aforementioned approach. We outline the degradations performed on ground-truth frames to obtain corresponding noisy frames used as surrogates for upsampled low-res frames in the Python function \texttt{ process$\_$image$\_$resize$\_$noise$\_$blur} presented in Listing \ref{code:degrad}. The code elucidates the procedure for performing the degradations necessary for learning to super-resolve frames at $128\times128$ resolution to $256\times256$ resolution.

\subsection{Quantitative results for image sequence generation on SkyTimelapse}
\label{app_sec: sky_quant}
We provide quantitative metrics for 16 and 128 length video generation on the SkyTimelapse dataset \citep{skydset} in Table \ref{tab:main-quant-sky}. This is in supplement to Figure \ref{fig: main-quant} (a), wherein the metrics for certain poorly performing methods might be difficult to elicit from the given plot due to a fine scale on the y-axis. As is evident, our method comprehensively outperforms the SoTA on the task.

\subsection{Statement of Broader Impact}
\label{app_sec: impacts}
Our work learns to model image sequences in a generative setting. Therefore, it does entail the risk of being misused like any other photorealistic image or video generative model. Therefore, its authentic distribution and ethical usage are essential. We shall release our model through GitHub or Huggingface. Both of which follow best practices to maintain community standards for ethical usage. We shall also include a widely accepted license in our release to prevent irresponsible usage. We would like to remind the reader that we only claim that our synthetic 3D CT Volumes bear statistical and visual resemblance to the Volumes present in the CT-RATE dataset curated by \citet{ct-rate-1}. Consequently, their real-world medical utility is yet to be established. As a result, users should refrain from using our work for real-world healthcare applications unless approved by appropriate medical authorities. We defer evaluation of our results from a medical standpoint to future follow-up work. The fact that our method uses datasets wherein it is hard for any bias to creep in comforts us in the quality of our work. Moreover, our work could advance image sequence generation in unconventional fields as well, leading to newfound applications in different domains of science and society.

\subsection{Validating our approach's unique positioning}
\label{app_sec:positioning}
Pixel-space data representation reformulation has been a widely overlooked aspect in image-sequence generation design space, almost as if it were \textit{hidden in plain sight}. In that context, GriDiT is uniquely positioned by being the first and only approach to capitalize on this aspect to address the synthesis quality versus efficiency tradeoff. We devote this section to taking a proof-by-contradiction approach to further validate our thesis. Specifically, we construct two naive baselines using the DiT using conventional data representations and contrast our performance with them. We present our experimental results in this regard in Table \ref{tab:positioning}.
\paragraph{Naive Baseline I: Channel-wise stacking paired with DiT.} 
\begin{table}
    \centering
    \caption{Additional Comparison with naive baselines to validate our unique positioning in the design space. \textbf{Bold:} best.}
    \label{tab:positioning}
\begin{tabular}{|lllll|}
\hline
\textbf{Method}                                                                                          & \textbf{Dataset}                                                                                        & \textbf{FVD-16 $\downarrow$}                                              & \textbf{FVD-128 $\downarrow$}                                              & \textbf{FVD-256 $\downarrow$}                                              \\ \hline
\multicolumn{1}{|l|}{\begin{tabular}[c]{@{}l@{}}Naive Baseline I\\ (Channel-wise Stacking)\end{tabular}} & \multicolumn{1}{l|}{}                                                                                   & \cellcolor[HTML]{FFFDFA}{\color[HTML]{333333} 85.5}          & \cellcolor[HTML]{FFFDFA}{\color[HTML]{333333} 313.2}          & \cellcolor[HTML]{FFFDFA}{\color[HTML]{333333} 452.5}          \\
\multicolumn{1}{|l|}{\cellcolor[HTML]{C0C0C0}\textbf{Ours}}                                              & \multicolumn{1}{l|}{\multirow{-2}{*}{\begin{tabular}[c]{@{}l@{}}SkyTimelapse\\ (256x256)\end{tabular}}} & \cellcolor[HTML]{C0C0C0}{\color[HTML]{2C3A4A} \textbf{64.1}} & \cellcolor[HTML]{C0C0C0}{\color[HTML]{2C3A4A} \textbf{183.4}} & \cellcolor[HTML]{C0C0C0}{\color[HTML]{2C3A4A} \textbf{206.6}} \\ \hline
\multicolumn{1}{|l|}{\begin{tabular}[c]{@{}l@{}}Naive Baseline II\\ (3D VAE)\end{tabular}}               & \multicolumn{1}{l|}{}                                                                                   & 129.32                                                       & -                                                             & -                                                             \\
\multicolumn{1}{|l|}{\cellcolor[HTML]{C0C0C0}Ours}                                                       & \multicolumn{1}{l|}{\multirow{-2}{*}{\begin{tabular}[c]{@{}l@{}}Taichi\\ (256x256)\end{tabular}}}       & \cellcolor[HTML]{C0C0C0}\textbf{118.919}                     & \cellcolor[HTML]{C0C0C0}-                                     & \cellcolor[HTML]{C0C0C0}-                                     \\ \hline
\end{tabular}
\end{table}
We establish this baseline by simply using a vanilla representation comprising channel-wise stacked frames for each training datapoint. In essence, we treat them as tensors bearing shape $f\times\frac{H}{K}\times\frac{W}{K}$ wherein $f$ represents the number of frames. We use the DiT's 2D VAE to embed each frame into the latent space sequentially and stack those latents channel-wise. Subsequently, we modify the DiT's projection layers to work with the inflated channel dimensions and train the model with the same recipe as our model. At inference, we use an Autoregressive sampling algorithm similar to ours in every aspect except for using the last $\frac{f}{K}$ channels as the sampling control signal. Finally, we use the same stage-2 (\emoticon{images/2.png}) as ours to ensure fairness of comparison. 
\paragraph{Naive Baseline II: 3D VAE paired with DiT.}  In this case, we replace the DiT's 2D VAE with the 3D VAE used by CogVideoX \citep{cogvideox}. The modification allows us to directly embed a sequence tensor bearing shape $f\times\frac{H}{K}\times\frac{W}{K}$ to a single latent that can be used to train the DiT model. As with the other experiment,  we pair this modified model with our stage-2 (\emoticon{images/2.png}) model for performing SR. Since this baseline cannot provide a structured sampling control signal, we restrict our experimentation in this setup to 16-length sequences only. We made sure to only use chose to use the Taichi dataset for these experiments because the 3D VAE yielded highly accurate reconstruction for those videos quantified by a reconstruction PSNR of 27.58 dB. 

Given the experiments in Table \ref{tab:positioning}, we make the three key inferences. First, our approach outperforms these baselines convincingly, thereby establishing the non-trivial nature of our contributions. Second, the benefits of our approach do not stem entirely from the inductive biases captured by the DiT-based diffusion paradigm; rather, all our design elements come together nicely to achieve our performance metrics. Third, the grid-based formulation lends itself better to diffusion-inpainting inspired autoregressive generation than a vanilla channel-wise stacked representation, making it crucial for utilizing the self-attention prior learned by the  DiT in our method. In essence, these experiments underscore the need for looking beyond conventional modeling approaches and focusing on devising better data representations in the domain.
\lstdefinestyle{python}{
    language=Python,
    backgroundcolor=\color{codebg},
    basicstyle=\ttfamily\small,
    keywordstyle=\color{defcolor}\bfseries,
    commentstyle=\color{commentcolor}\itshape,
    stringstyle=\color{stringcolor},
    showstringspaces=false,
    breaklines=true,
    frame=single,
    rulecolor=\color{gray},
    numbers=none,
    keepspaces=true,
    columns=flexible,
}

\begin{lstlisting}[caption={Python code for performing the degradations required to train our stage-2 (SR) model.}, label={code:degrad}]
import os
import random
import cv2
import numpy as np

def add_gaussian_noise(image, mean=0, std=10):
    """Add Gaussian noise to an image."""
    noise = np.random.normal(mean, std, image.shape)
    noise = noise.astype(np.float32)
    noisy = cv2.add(image.astype(np.float32), noise)
    return np.clip(noisy, 0, 255).astype(np.uint8)

def process_image_resize_noise_blur(
    image_path,
    erosion_iterations=3,
    blur_radius=15,
    brightest_fraction=0.4,
    global_blur_radius=7):
    """
    Load an image, resize, add noise, and apply Gaussian blur
    with random parameters.
    """
    # Load and convert image to RGB
    image = cv2.imread(image_path)
    image = cv2.cvtColor(image, cv2.COLOR_BGR2RGB)
    # Resize image twice: 256x256 via 128x128
    image_resized = cv2.resize(
        cv2.resize(image, (128, 128), interpolation=cv2.INTER_CUBIC),
        (256, 256), interpolation=cv2.INTER_CUBIC
    )
    # Add Gaussian noise with random std
    std = random.randint(10, 15)
    noisy_image = add_gaussian_noise(image_resized, std=std)
    # Randomly apply Gaussian blur
    if random.random() < 0.5:
        result_image = noisy_image
    else:
        blur_sizes = [9, 11, 13, 15]
        blur_radius = random.choice(blur_sizes)
        result_image = cv2.GaussianBlur(
            noisy_image, (blur_radius, blur_radius), 0
        )

    return result_image

\end{lstlisting}


\section{Additional qualitative results}
\label{appsec:additional_qual}
We present and analyze additional qualitative results to bring forth a better understanding of the pertinent aspects of our method.  

\subsection{Arbitrary length video synthesis}
\begin{figure}[ht]
    \centering
    \includegraphics[width=\textwidth]{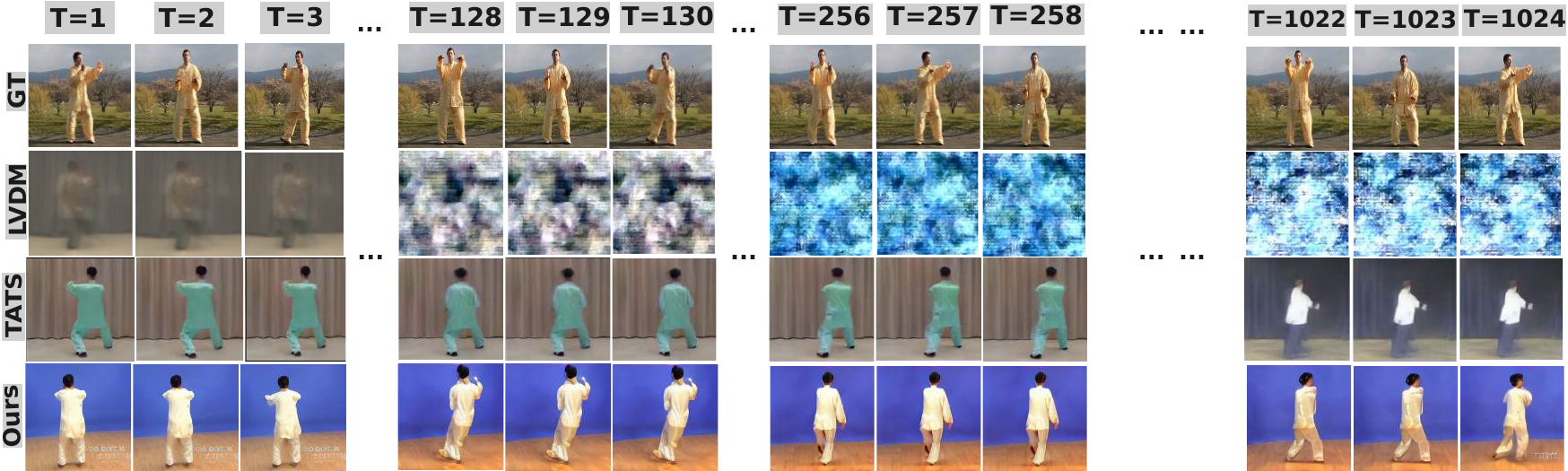}
    \caption{Qualitative comparisons for arbitrary length video synthesis at $256\times256$ resolution on the Taichi dataset. As presented in our Quantitative results in section 4.1 of the paper, our method offers: \textbf{(1) Superior perceptual quality} than the prior art, exemplified by the sharpness of details in our synthetic frames. \textbf{(2) Superior spatiotemporal consistency} than the prior art. There are no changes in the overall `scene' or `random jumps' even when the synthesis is extended to arbitrary lengths. \textbf{(3) Freedom from leakage and memorization issues} as it generates arbitrarily long videos despite being trained only on videos with $\le$ 300 frames. ($T:$ Frame indices. GT: Ground Truth. Please note that the ground truth video has been looped repeatedly to display 1024 frames, despite being only 293 frames long.)} 
    \label{fig: long_qual_taichi}
\end{figure}

We tie this discussion to the results presented in section \ref{subsec: results} of the paper. The Taichi dataset \citep{taichi_dset} is a particularly challenging dataset from a video generation standpoint because it requires a model to infer large motion and high-frequency details from very few data points at a relatively low resolution ($256\times256$). Consequently, most prior methods in the domain struggle to get both motion and high-frequency details right on this dataset. In Figure \ref{fig: long_qual_taichi}, we compare our method's performance with the SoTA on the Taichi dataset \citep{taichi_dset}, qualitatively. Therein, we make three key inferences. First, our method yields superior perceptual quality than both LVDM \citep{lvdm} and TATS \citep{tats} on arbitrarily long generation. We attribute this to the efficacy of our Grid-based Autoregressive sampling algorithm.
We also notice a severe decline in quality with increasing sequence length in our competing methods. Such a decline indicates the overall worse modeling capability of approaches that seek to `extrapolate' long videos from a few synthetic frames without applying appropriate inductive biases to the process. Second, we also outperform these methods in terms of long-range temporal consistency. The consistency is evident from the fact that our `overall scene' remains the same throughout the 1024 frames. Whereas it gets destroyed or changes for the other methods. We attribute the improvement in consistency to DiT's \citep{dit} strong self-attention prior, which provides inductive biases to our sampling algorithm. Third, our ability to synthesize much longer videos than the ones seen in training asserts our remediation of the widely prevalent leakage \citep{leakage} and memorization \citep{memorization} issues in generative models.  

\begin{figure}[ht]
    \centering
    \includegraphics[width=\textwidth]{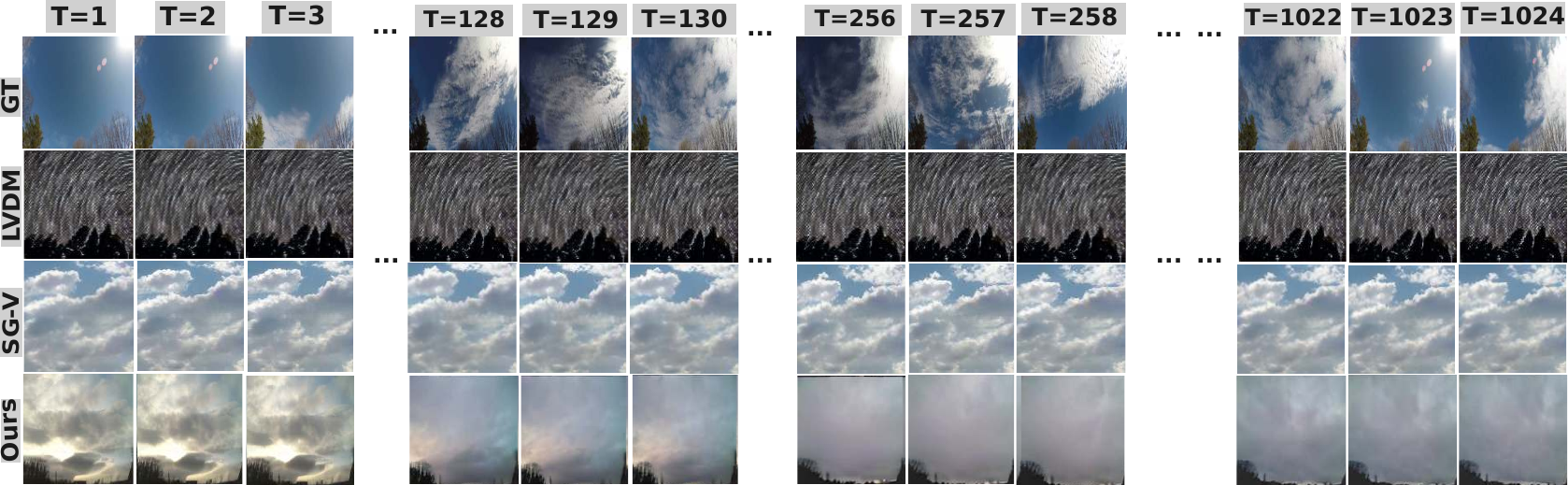}
    \caption{We compare our method with the SoTA for arbitrary length video synthesis at $256\times256$ resolution on the SkyTimelapse dataset qualitatively. Consequently, we observe that: \textbf{(1)} LVDM struggles in terms of photorealism of the synthetic frames and modeling motion across them. Moreover, it mostly generates scenes in low light with limited variability. \textbf{(2)} StyleGAN-V yields comparable per-frame quality but struggles in terms of variation in the scene lighting and motion across frames. We observed the `looping artifacts' that we specifically mitigate by using 3D positional embeddings in our method. \textbf{(3)} Whereas our method generates high-quality spatiotemporally coherent videos with substantial diversity in lighting across different frames. ($T:$ Frame indices. GT: Ground Truth. SG-V: StyleGAN-V. Please note that the ground truth video has been looped repeatedly to show 1024 frames despite it being only 361 frames long.)}
    \label{fig: long_qual_sky}
\end{figure}

Figure \ref{fig: long_qual_sky} illustrates a similar analysis for the SkyTimelapse dataset \citep{skydset}. Although the aforementioned benefits of our method are evident here as well, a few more interesting ones emerge. Specifically, we observe that LVDM \citep{lvdm} collapses to a mode of dark scenes across various iterations of unconditional sampling. Whereas,  StyleGAN-V \citep{style-gan-v} performs comparably to our method in terms of per-frame quality. Yet, it is worse in terms of FVD due to the presence of `looping artifacts', which we resolve in our method using 3D positional embeddings. Moreover, both these methods struggle in terms of variability. LVDM struggles with viewing angle and lighting variations across and within its synthetic videos. However, StyleGAN-V struggles only in per-frame variability within different synthetic videos. Our method performs better on both of these fronts. \subsection{Different sampling settings}
\begin{figure}[ht]
    \centering
    \includegraphics[width=\textwidth]{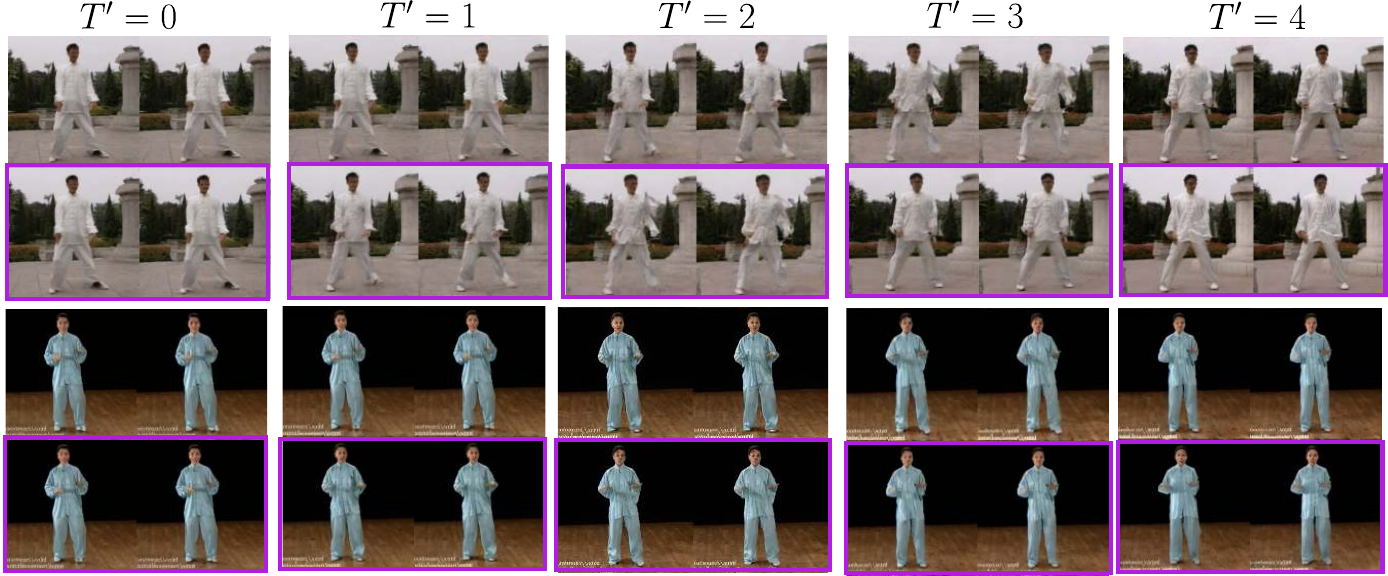}
    \caption{We perform Grid-based Autoregressive sampling for arbitrary length video generation in the $\{K=2$, one-row control signal$\}$ setting on the Taichi dataset. We observe that the setting sacrifices spatiotemporal consistency for slight gains in per-frame quality. (\emoticon{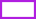}: Sampling control signal.)}
    \label{fig: grid-2}
\end{figure}

\begin{figure}[h]
    \centering
    \includegraphics[width=\textwidth]{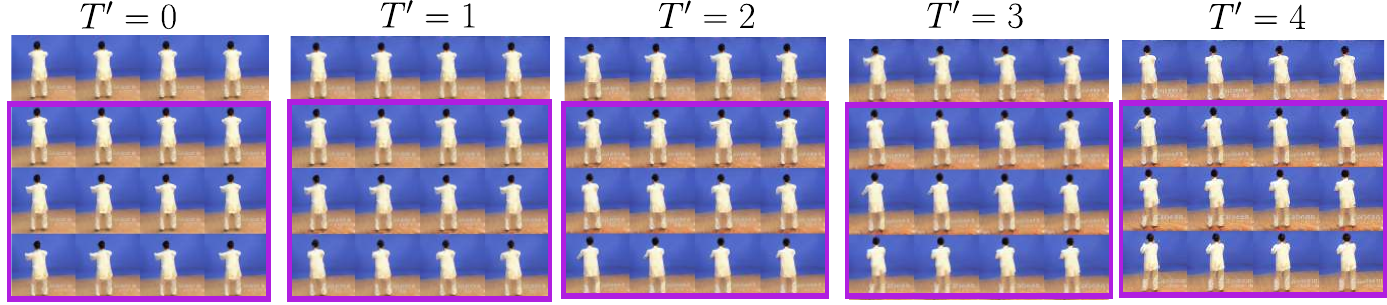}
    \caption{We perform Grid-based Autoregressive sampling for arbitrary length video generation in the $\{K=4$, three-row control signal$\}$ setting on the Taichi dataset. This setting is the sweet spot of the tradeoff between per-frame quality and long-range temporal consistency. (\emoticon{images_supplementary/violet-box.png}: Sampling control signal.)}
    \label{fig: grid-4}
\end{figure}

\begin{figure}[ht]
    \centering
    \includegraphics[width=\textwidth]{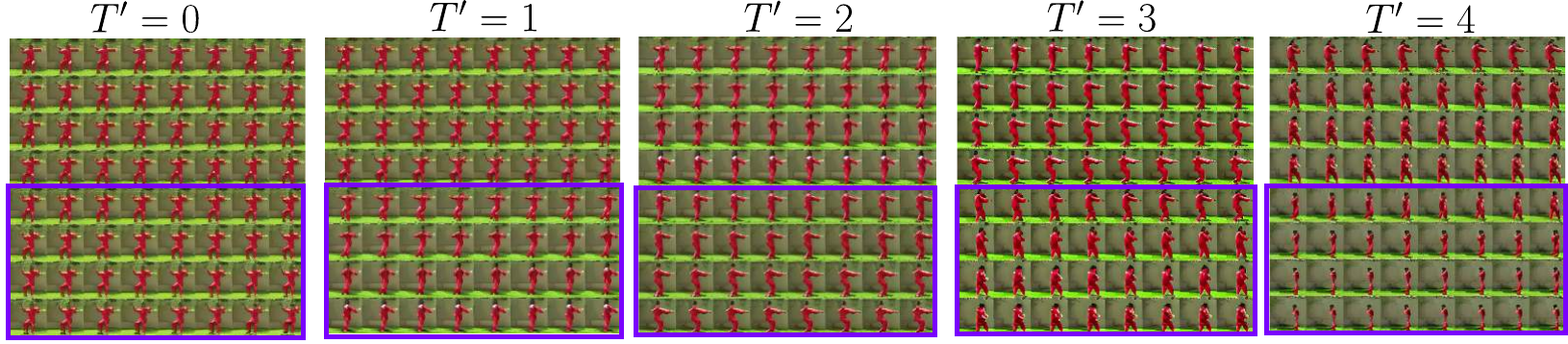}
    \caption{We perform Grid-based Autoregressive sampling for arbitrary length video generation in the $\{K=8$, four-row control signal$\}$ setting on the Taichi dataset. We observe that the setting sacrifices per-frame quality for slight gains in spatiotemporal consistency. (\emoticon{images_supplementary/violet-box.png}: Sampling control signal.)}
    \label{fig: grid-8}
\end{figure}

We dedicate this section to examining the interplay between the various variables associated with our Grid-based autoregressive sampling scheme. In figures \ref{fig: grid-2}, \ref{fig: grid-4}, and \ref{fig: grid-8} we present the first five iterations of our Grid-based Autoregressive sampling algorithm in its $\{K=2$, one-row control signal$\}$, $\{K=4$, three-row control signal$\}$, and $\{K=8$, four-row control signal$\}$ settings on the Taichi dataset, respectively for step 1 of our sampling scheme. In all these experiments, we interpolated $K/2$ grid elements in step 2 of the sampling scheme. The figures conform to the findings of section \ref{subsec: ablations} of the paper, wherein we observe a tradeoff between the amount of temporal signal and spatial details a setting has to offer.
\clearpage
\clearpage
\begin{figure}[h]
\centering
\includegraphics[width=\textwidth]{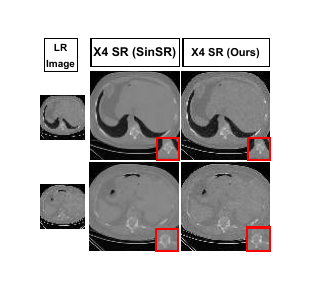}
    \caption{Qualitative comparison for the SR task on the CT-RATE dataset. Our method's stage 2 (\emoticon{images/2.png}) qualitatively outperforms  SinSR (finetuned) on the super-resolution and fine information addition task, especially in high-frequency regions, as highlighted by the inset.}
    \label{fig: sr-compar-1}
\end{figure}
\clearpage
\begin{figure}[h]
    \centering
    \includegraphics[width=\textwidth]{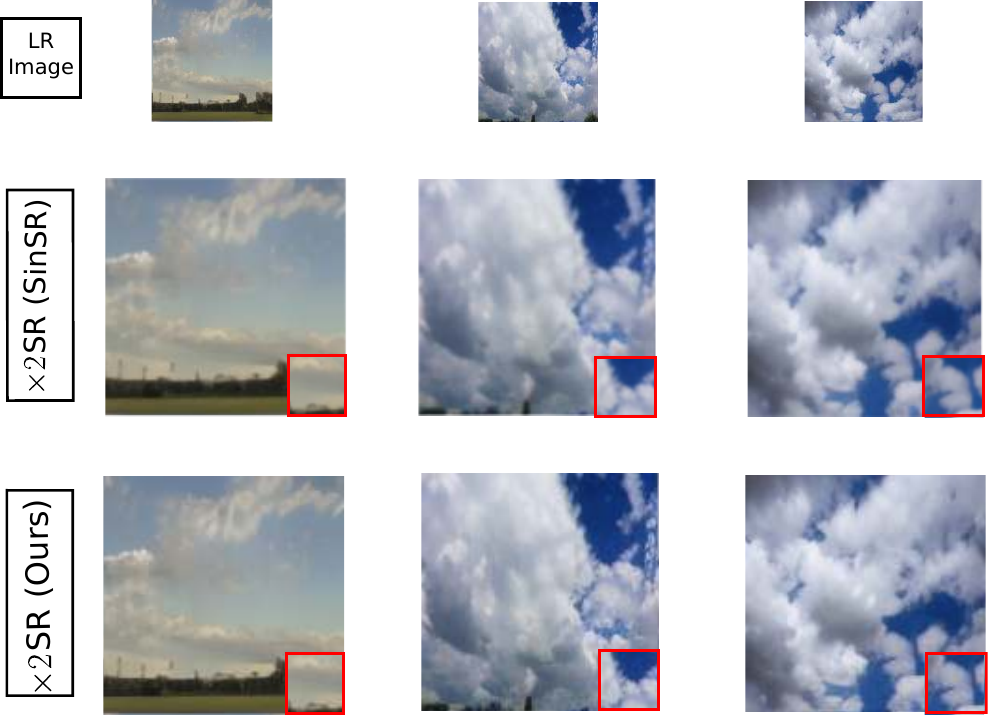}
    \caption{Qualitative comparison for the SR task on the SkyTimelapse dataset. Our method's stage 2 (\emoticon{images/2.png}) qualitatively outperforms  SinSR in refining individual frames, as highlighted by the inset.}
    \label{fig: sr-compar-2}
\end{figure}

\clearpage
\end{document}


\maketitle
\input{sections_supplementary/main_supplementary}
\input{sections_supplementary/qualitative_results}
\clearpage
\bibliographystyle{IEEEtran}
\bibliography{Styles/refs}